\documentclass{article}

\PassOptionsToPackage{numbers, sort}{natbib}
 \usepackage[preprint]{neurips_2026}

\usepackage[utf8]{inputenc} 
\usepackage[T1]{fontenc}    
\usepackage[hyperfootnotes=false]{hyperref}       
\usepackage{url}            
\usepackage{booktabs}       
\usepackage{amsfonts}       
\usepackage{nicefrac}       
\usepackage{microtype}      
\usepackage{xcolor}         
\usepackage{graphicx}
\usepackage{amsmath}
\usepackage{afterpage}
\usepackage[para]{footmisc}
\usepackage{wrapfig}
\usepackage{authblk}
\usepackage[font=small,labelfont=bf]{caption}

\usepackage{array,booktabs,multirow,xltabular}

\usepackage{longtable}
\usepackage{array}
\usepackage{tabularx}
\usepackage{makecell}
\usepackage{multirow}
\usepackage{subcaption}
\usepackage{graphicx}

\usepackage{ragged2e}
\newcolumntype{L}[1]{>{\RaggedRight\arraybackslash}p{#1}}

\usepackage{cleveref}
\crefname{equation}{Eq.}{Eqs.}
\Crefname{equation}{Equation}{Equations}
\crefname{section}{Sec.}{Secs.}
\Crefname{section}{Section}{Sections}
\crefname{subsection}{Sec.}{Secs.}
\Crefname{subsection}{Section}{Sections}
\crefname{figure}{Fig.}{Figs.}
\Crefname{figure}{Figure}{Figures}
\crefname{table}{Tab.}{Tabs.}
\Crefname{table}{Table}{Tables}

\usepackage{titletoc}

\newcommand{\movebench}{\textsc{MoveBench}}

\title{\movebench: A Benchmark for\\Global-Scale Wildlife Movement Forecasting}

\author{%
\textbf{Justin Kay}$^{1}$\thanks{Correspondence to \texttt{kayj@mit.edu} \qquad \qquad \qquad \quad \quad $^{\alpha, \beta}$ Equal contribution within groups, alphabetically ordered} \qquad \qquad 
\textbf{Shir Bar}$^{1}$ \qquad \qquad 
\textbf{Ellen O. Aikens}$^{2,\alpha}$ \qquad \qquad \textbf{Martin Becker}$^{3,\alpha}$ \quad 
\textbf{Francesca Cagnacci$^{4,\alpha}$} \quad \textbf{Juliet Cohen$^{5,\alpha}$} \quad \textbf{Scott W. Forrest$^{6,\alpha}$} \quad \textbf{Jessica Kendall-Bar$^{7,\alpha}$} \quad
\textbf{Madeleine Lucas$^{8,\alpha}$} \quad 
\textbf{Macon Overcast$^{18,\alpha}$} \quad
\textbf{Meredith S. Palmer$^{9,\alpha}$} \quad 
\textbf{Will Rogers$^{9,\alpha}$} \quad \textbf{Nicholas J. Russo$^{10,\alpha}$} \quad
\textbf{Christian Rutz$^{11,\alpha}$} \quad \textbf{Larissa T. Beumer$^{12,\beta}$} \quad 
\textbf{Michael Brown$^{20,\beta}$} \quad 
\textbf{Ying-Chi Chan$^{13,\beta}$} \quad  \textbf{Sarah C. Davidson$^{14,\beta}$} \quad
\textbf{Diego Ellis Soto$^{15,\beta}$} \quad \textbf{Anne G. Hertel$^{16,\beta}$} \quad \textbf{Roland Kays$^{17,\beta}$} \quad \textbf{Benjamin Koger$^{2,\beta}$}  \quad
\textbf{Guram Mikaberidze$^{2,\beta}$} \quad \textbf{Thomas Mueller$^{16,\beta}$} \quad \textbf{Ruth Oliver$^{5,\beta}$} \quad \textbf{Thorsten Papenbrock$^{3,\beta}$} \quad
\textbf{Robert Patchett$^{11,\beta}$} \quad \textbf{Jared A. Stabach$^{18,\beta}$} \quad \textbf{Dane Taylor$^{2,\beta}$} \quad \textbf{Scott W. Yanco$^{19,\beta}$} \quad 
\textbf{Sara~Beery$^{1}$}\\
$^1$MIT \quad 
$^2$University of Wyoming \quad 
$^3$Marburg University, Hessian.AI \quad 
$^4$Fondazione Edmund Mach, Research and Innovation Centre, Animal Ecology Unit\quad
$^5$UCSB \quad 
$^6$Queensland University of Technology \quad 
$^7$Scripps Institution of Oceanography, UCSD \quad
$^8$Colorado State University \quad 
$^9$Yale \quad 
$^{10}$Harvard \quad 
$^{11}$University of St. Andrews \quad 
$^{12}$UNIS \quad 
$^{13}$NTU Singapore \quad 
$^{14}$Max Planck Institute of Animal Behavior \quad
$^{15}$UC~Berkeley \quad 
$^{16}$Senckenberg Biodiversity and Climate Research Centre \quad 
$^{17}$NCSU and NC Museum of Natural Sciences \quad
$^{18}$Smithsonian NCBI \\
$^{19}$Smithsonian Institution \quad
$^{20}$Giraffe Conservation Foundation
}

\begin{document}

\maketitle

\begin{abstract}
  Understanding and predicting wildlife movement is critical for ecology and conservation. While trajectory forecasting has advanced for human and vehicle movement, wildlife trajectories present distinct challenges: they are unconstrained in space, highly stochastic, and influenced by 
  environmental conditions. We introduce \movebench, the first large-scale benchmark for probabilistic wildlife movement forecasting, containing 2.6M GPS locations from 800+ individuals across 110 species in 127 countries, paired with 1.6B environmental raster tiles capturing 160 covariates known or hypothesized to influence movement. We propose a probabilistic evaluation protocol for movement trajectory forecasts, addressing limitations of point-prediction metrics for inherently stochastic phenomena. Through comprehensive empirical evaluation of four method families across multiple temporal and spatial scales, we reveal that: (1)~existing predictive methods generalize better to future timepoints than to unseen individuals, 
  (2)~deep learning approaches do not consistently outperform simpler baselines, and 
  (3)~environmental covariate selection significantly impacts performance. 
  \movebench\ enables standardized evaluation of movement forecasting methods and provides a foundation for methodological advances on this
  ecologically important task. \linebreak[3]
  \end{abstract}

\section{Introduction}

\afterpage{\clearpage} 
\begin{figure}[p]
    \centering
    \includegraphics[width=\linewidth]{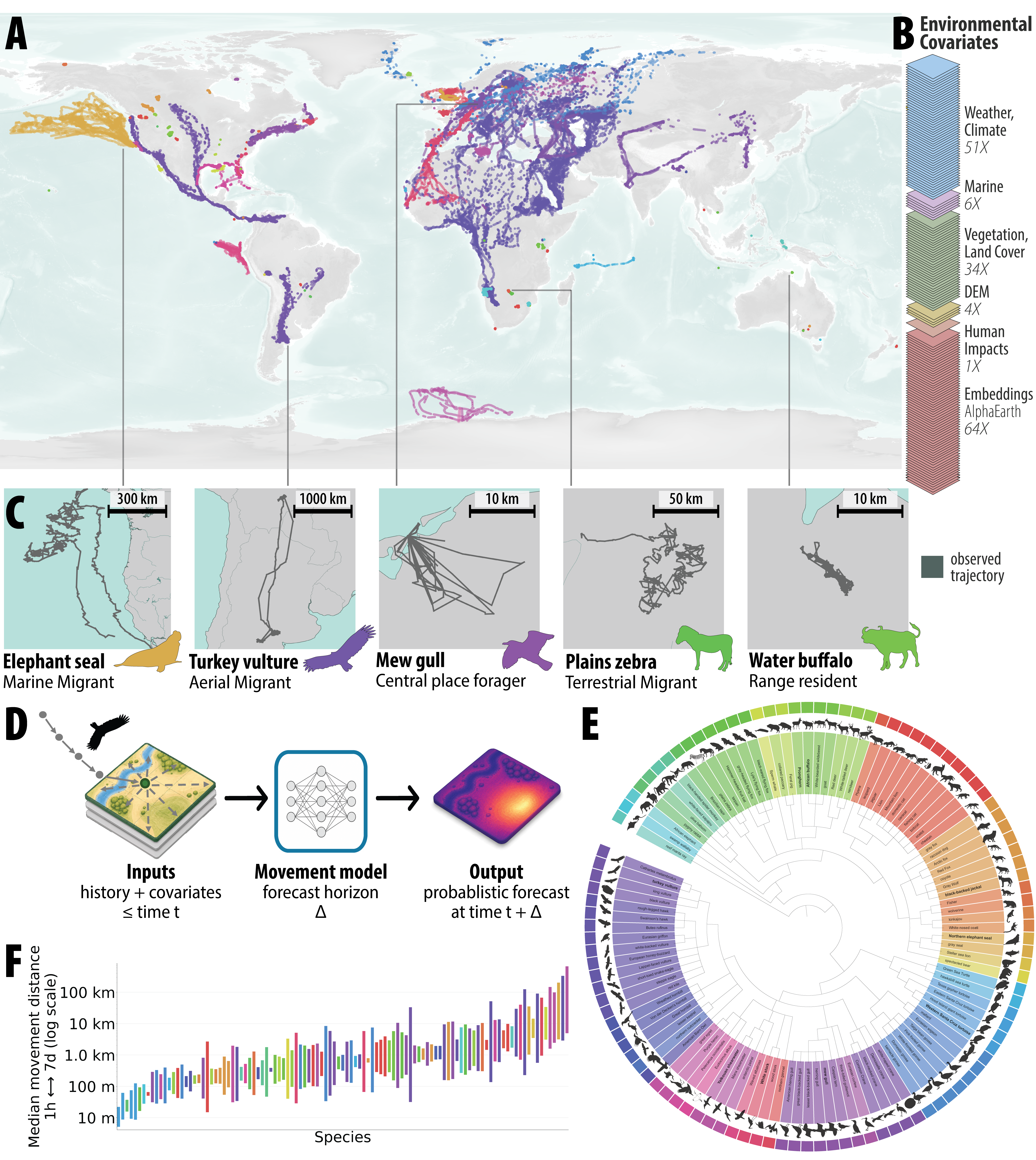}
    \caption{\small
    \textbf{\movebench\ is the first global benchmark for animal movement forecasting.} 
    \textbf{(A)}~The benchmark contains 2.6 million curated GPS locations from over 800 individuals of 110 wildlife species, with observations from 127 countries spanning all continents, and \textbf{(B)}~1.6 billion raster tiles representing 160 associated environmental variables known or hypothesized to influence movement.
    \textbf{(C)}~Five example species are shown that illustrate the diversity of wildlife movement: from northern elephant seals and turkey vultures that migrate thousands of kilometers over land and sea, to plains zebras that migrate over land, to mew gulls and water buffalo that spend most of their lives centered around one location.
    \textbf{(D)}~The predictive challenge is \textit{probabilistic forecasting}: models map from an individual's movement history, environmental conditions, and other covariates to a probability distribution over possible future locations of that individual at a specified time horizon.
    \textbf{(E)}~The data in \movebench\ covers 110 species across the tree of life with \textbf{(F)}~movement patterns spanning over five orders of magnitude in spatial scale. Bars in (F) extend vertically from the median 1-hour movement distance to the median 7-day movement distance of each species. Trajectories in (A), silhouettes in (C), and bars in (F) correspond by color to species in (E).
    }
    \label{fig:hero}
\end{figure}

Movement is a fundamental process of life on Earth and is at the basis of key ecological functions, from pollination and nutrient cycling to disease transmission and carbon sequestration~\cite{nathan2008movement,jeltsch2013integrating,siegel2023quantifying}.
Advancements in animal tracking technology 
(\textit{e.g.,} GPS-based radio-telemetry~\cite{cagnacci2010animal})
combined with real-time data transmission have allowed scientists to track and record wildlife movement at unprecedented scales and resolutions and for an increasingly diverse array of species~\cite{kays2015terrestrial}. This has catalyzed an explosion in animal movement data collection over the past decade, as evidenced by the over ten billion logged animal locations on the animal tracking data repository Movebank~\cite{web205Movebank}.
This breadth of observational data has deepened our understanding of how and why animals move through the world~\cite{aikens2022viewing, mouritsen2018long,jesmer2018ungulate}, 
helped prioritize areas for protection~\cite{blondin2020combining,morrison2015space,hindell2020tracking}, and informed strategies for mitigating human-wildlife conflict~\cite{graham2009movement,blondin2020combining,block2011tracking,pekarsky2021using}.
Beyond observational insights, the ability to predict how future animal movement might respond to various changes in natural and built environments---for instance, how new roads or energy infrastructure may alter movement corridors, how changing land use may reshape landscape connectivity~\cite{cowan2025impact, dorber2023new,Hofmann2023connect}, and how migratory routes may shift under climate change~\cite{hofmann2024dispersal}---could revolutionize conservation decision making~\cite{gomez2025understanding,singh2011conserving,doherty2021human,gaynor2024anthropogenic,ellis2023vision,verzuh2026aligning}.

Currently, we do not have the ability to robustly 
forecast animal movement. 
While many methods have been proposed for analyzing animal movement data~\cite{fortin2005wolves,avgar2016integrated,munden2021did,nicosia2017multi,klappstein2023flexible,bakner2023behavioral,klappstein2024step,muff2020accounting,potts2022assessing, schlagel2019estimating,chatterjee2024modelling,eisaguirre2024rayleigh,hofmann2024methods,valle2024bridging,egan2025accounting,forrest2025predicting}, their predominant use in the literature is for 
description, inference, and retrospective analysis rather than prediction~\cite{hacker2026making}. 
There is great potential for the field of movement ecology to move towards a predictive science, and growing interest in the development of predictive methods~\cite{forrest2026predicting,forrest2025predicting,potts2022assessing,signer2024simulating,cifka2023moveformer, ranc2022memory,hacker2026making}.
To realize this potential, the field needs clear and shared definitions of success.
This requires benchmarking frameworks that evaluate models across the dimensions that matter for real-world use: different species, movement modes (\textit{e.g.,} walking, swimming, flying), spatial and temporal scales, and environmental contexts. 
No such benchmark exists for animal movement forecasting.

Crucially, it is not possible to simply transfer lessons from other trajectory prediction benchmarks onto animal movement data. The most common trajectory data sources in the machine learning literature---human and vehicle trajectories---are fundamentally different, and in many ways simpler, than wildlife trajectories.
For instance, they are almost always map-aligned,
\textit{i.e.}, explicitly map-matched to known linear transportation infrastructures supporting human mobility,
allowing movement to be modeled as a simplified process evolving over a constrained set of 
paths \cite{newson2009hidden,wang2022ltp}. There are no such maps 
for animal movement---in fact, discovering what routes are taken through a landscape is often an explicit goal of collecting the data in the first place~\cite{nathan2008movement,thurfjell2014applications,northrup2022conceptual}. 
Further, existing trajectory forecasting benchmarks primarily evaluate only a single best-guess future trajectory with metrics such as average displacement error, rather than probabilistic forecasts representing the range of possible future trajectories. Animal movement is highly stochastic~\cite{patterson2008state}, with several plausible futures at any given point in time, so accurate prediction requires both probabilistic forecasts and metrics that can assess the distribution of plausible futures.
Finally, animal movement is heavily influenced by 
local environmental conditions~\cite{nathan2008movement,northrup2022conceptual,patterson2008state}, and a key goal of forecasting is to predict movement under changing environments. 
Incorporating this environmental context into predictive methods requires synthesizing heterogeneous, multi-scale, multi-resolution, and time-varying data sources. 
This relationship to covariates is a complexity missing from existing human and vehicle trajectory methods, and
addressing these challenges will require significant methodological innovation. 

We introduce \movebench, the first large-scale benchmark dataset for animal movement forecasting. The dataset contains 2.6 million curated GPS locations from over 800 individuals of 110 wildlife species, with observations from 127 countries spanning all continents. Alongside the GPS trajectory data, \movebench\ incorporates 160 publicly-available and global environmental covariates, which are preprocessed into 1.6 billion environmental raster tiles comprised of over two trillion environmental measurements. We propose a set of metrics for evaluating probabilistic movement forecasting methods and conduct the first large-scale empirical comparison of predictive movement methods for this task. Our experiments demonstrate there is much room for improvement in probabilistic forecasting methods, particularly in out-of-domain generalization, covariate selection, and multiscale spatiotemporal reasoning.
Altogether, \movebench~represents both the largest dataset for animal movement forecasting in existence and the most extensive empirical comparison of animal movement prediction methods ever conducted, while providing a novel environmentally-conditioned probabilistic spatiotemporal forecasting challenge to the ML community.
The benchmark and codebase will be made publicly available,
enabling future research and methods development.\footnote{Paper currently under review. The benchmark will be officially released following acceptance, at which time this preprint will be updated with links to code and data.}

\vspace{-5pt}
\section{Related work}
\label{sec:related}
\vspace{-5pt}

Compared to existing benchmarks for trajectory prediction (see \cref{tab:datasets}), \movebench\ offers innovations along three axes (for additional related work, see Appendix \cref{appendix:related}): 

\textbf{Data:} The wildlife movement data in \movebench\ is more complex, heterogeneous, and covariate-dependent than existing benchmarks focusing on human movement~\cite{pkdd-15-predict-taxi-service-trajectory-i,liu2023graphmm,zhu2024unitraj,yang2016participatory,cho2011friendship,li2024limp,zheng2010geolife}. It is unconstrained in space, varies drastically in pattern and scale across species and geographies, and requires synthesizing large-scale geospatial datasets in order to provide methods with the environmental context needed to predict where animals will move~\cite{thurfjell2014applications,northrup2022conceptual}. 
Existing large datasets with tracked animals do not support probabilistic movement forecasting---BEBE~\cite{hoffman2024bebe} is a benchmark for behavior classification in accelerometry data, and \citet{noonan2019comprehensive} evaluates methods for space use estimation over a large collection of datasets, anonymizing both GPS locations and timestamps.

\textbf{Task:} \movebench\ focuses on probabilistic forecasting to reflect the inherently stochastic nature of animal movement (see \cref{sec:setup}). Prior work primarily evaluates forecasting based on either: (i)~point-prediction metrics (\textit{e.g.}, L2 error between ground truth and predicted locations), which do not assess the quality of entire predictive distributions ~\cite{shi2021sgcn,shi2023tutr,pellegrini2009you}, (ii)~qualitative inspection~\cite{signer2024simulating}, or (iii)~log loss~\cite{forrest2026predicting,cifka2023moveformer}, which is probabilistic but cannot be used to compare across arbitrary model classes because it requires an explicit predictive density, which is unavailable for deterministic predictors and may be intractable for some methods. 
As discussed in \cref{sec:metrics}, \movebench\ introduces an evaluation methodology based on a \textit{proper scoring rule}~\cite{gneiting2007strictly,waghmare2025proper} called the \textit{energy score} that enables robust evaluation of probabilistic forecasts from any trajectory prediction model~\cite{shahroudi2024evaluation}.

\textbf{Methods:} 
We compare commonly-used and state-of-the-art methods for predicting animal movement, which model the movement process as a function of \textit{habitat selection}, whereby an individual chooses where to go next based on environmental conditions at candidate future locations~\cite{fortin2005wolves,avgar2016integrated,forrest2026predicting,cifka2023moveformer} coupled with their estimated movement capacity based on past movement. \movebench\ enables empirical comparison at scale between these methods for the first time.



  \begin{table}[t]
    \centering
    \caption{\small
    \textbf{Trajectory benchmarks.} Prior large-scale trajectory benchmarks focus on predicting vehicle movement on roads. \movebench\ expands this to enable probabilistic evaluation on a continuous global spatial domain $\mathcal{X}$ rather than point forecasts on a constrained road map. \movebench\ also includes more classes with diverse movement patterns, more covariates that influence movement, and larger spatial and temporal ranges than prior work.
    \textit{\footnotesize IQR computed for per-trajectory cumulative distances and time ranges.
    }
    }
    \label{tab:datasets}
    \resizebox{\columnwidth}{!}{
    \begin{tabular}{c|c|c|c|c|c|c|c|c}
    \toprule
        Benchmark & \# GPS & \# Cls & \# Cov & Dist IQR (km) & Time IQR (h) & Task & $\mathcal{X}$ & Probabilistic? \\
    \midrule
         \multirow{2}{*}{GeoLife~\cite{zheng2010geolife}} & \multirow{2}{*}{23.6M} & \multirow{2}{*}{10} & \multirow{2}{*}{0} & \multirow{2}{*}{3 -- 22} & \multirow{2}{*}{0.3 -- 3} & Forecast, & \multirow{2}{*}{Roads} &  \\
         & & & & & & Classify & & \\
         Porto~\cite{pkdd-15-predict-taxi-service-trajectory-i} & 83M & 1 & 4 & 2 -- 7 &  0.1 -- 0.2 & Forecast & Roads &  \\
         T-Drive~\cite{yuan2010t} & 790M & 1 & 0 & 860 -- 2000 & 145 -- 148 & Forecast & Roads &  \\
         WorldTrace~\cite{zhu2024unitraj} & 8.8B & 1 & 0 & 0.8 -- 10 & 0.03 -- 0.2  & Forecast & Roads &  \\
    
    
    \midrule
    \textbf{\movebench} & \textbf{2.6M} & \textbf{110} & \textbf{160} & \textbf{283 -- 3500} & \textbf{818 -- 8000} & \textbf{Forecast} & \textbf{Globe} & \textbf{\checkmark} \\

    \bottomrule

    \end{tabular}
    } 

\end{table}

\vspace{-6pt}
\section{Movement forecasting}
\label{sec:setup}
\vspace{-6pt}

The high-level goal is to predict the next step an animal will take at any time point, given its movement history and current environment.
The following precisely defines this movement forecasting task (see \cref{fig:hero}d), describing observed trajectories, forecast targets, and evaluation settings. 

\textbf{Trajectory observations.} We consider a spatiotemporal domain where \(\mathcal{X}\subseteq \mathbb{R}^2\) and \(\mathcal{T}\subseteq \mathbb{R}\) denote the spatial and temporal extents available to an animal, respectively.
For a single tracked animal, let
$\tau = \bigl((\ell_n,t_n)\bigr)_{n=1}^{N}$ denote an observed trajectory of length $N$, where \(\ell_n \in \mathcal{X}\) is the recorded location at observation time \(t_n \in \mathcal{T}\) for $t_1 < t_2 < \cdots < t_N$. For \(n=1,\dots,N-1\), define the elapsed time between consecutive observations by $\Delta t_n = t_{n+1}-t_n$.
We do not assume that \(\Delta t_n\) is constant in \(n\), although in many trajectories it is approximately regular.

\textbf{Covariates.} Each spatiotemporal point is associated with a set of covariates, \textit{i.e.} a feature vector. These may include movement-, time-, and environment-derived features. Formally, we represent these features by a vector-valued spatiotemporal field
$
c:\mathcal{X}\times\mathcal{T}\to\mathbb{R}^K,
$
where \(K\) is the number of features. For any location--time pair \((\ell,t)\in\mathcal{X}\times\mathcal{T}\), the vector \(c(\ell,t)\) represents corresponding features.
\(c\) 
can be evaluated for any observed point $c_n = c(\ell_n,t_n)$ with $n=1,\dots,N$, or at any candidate future location \(\tilde{\ell}\in\mathcal{X}\) and candidate future time \(\tilde{t}\in\mathcal{T}\). 
See \cref{tab:movebench_covariates_summary} for a summary of the covariates available in \movebench.

\textbf{Forecasting task.}
We evaluate probabilistic movement forecasting at forecast horizons \(\Delta>0\). For a forecast origin $(\ell, t)$, the prediction target at time horizon \(\Delta\) is the animal's location at time \(t+\Delta\), denoted $\ell_{t+\Delta}$. 
For an observed trajectory, define the history up to time \(t\) by
$\tau_{\le t} = \bigl((\ell_i,t_i)\bigr)_{i=1}^{n(t)}$ and any associated covariates by $c_{\le t} = \bigl(c(\ell_i,t_i)\bigr)_{i=1}^{n(t)}$, with $n(t) = \max\{i: t_i \le t\}$ representing the number of observed steps up to and including time $t$.

Then, for a given forecast horizon \(\Delta\) and origin time $t$, a movement forecasting method estimates a conditional predictive spatial distribution
$P_{t,\Delta}(\,\cdot \mid \tau_{\le t}, c_{\le t})$
over \(\mathcal{X}\) representing the probability the animal will move to any spatial location at time $t+\Delta$. 
In practice, for tractability, this is typically defined over a reasonable spatial range surrounding the animal's current location (\textit{e.g.}, its 99th percentile next-step movement distance) rather than all of \(\mathcal{X}\).
Deterministic predictors are a special case where \(P_{t,\Delta}\) is a point mass at a single predicted location.

\vspace{-6pt}
\section{\movebench}
\label{sec:movebench}
\vspace{-6pt}

\movebench\ (see \cref{fig:hero}) contains 2.6 million  GPS locations from over 800 individuals of 110 wildlife species, accompanied by over 1.6 billion environmental raster tiles. Here, we briefly describe how we select this data and construct the benchmark. More details can be found in Appendix \cref{sec:appendix-data}.

\textbf{Trajectory curation.}
We aggregate trajectory data from: (1) the Movebank Data Repository~\cite{movebankMovebank}, a subset of data from Movebank~\cite{web205Movebank} that have Creative Commons licensing, DOIs, and have already undergone some quality assurance and anomaly filtering, (2) a literature review of recent movement methods papers with empirical experiments on public data, and (3) a survey of domain experts in the Movement Biodiversity Observation Network~\cite{geobonMove}. 
This yields 147 raw trajectory datasets, from which we apply a coarse set of filtering criteria followed by manual curation (see Appendix \cref{sec:appendix-data} for details and source acknowledgments). 

\begin{table}[t]
\centering
\small
\setlength{\tabcolsep}{4pt}

\caption{
\small
\textbf{Covariates in \movebench.} The benchmark includes 160 environmental layers representing context at observed trajectories and at candidate locations for future movement. Covariates are included as rasters, centered at observed GPS points and aligned across spatial and temporal scales. Resolution indicates the spatial extent represented by one pixel. Cadence indicates the temporal  sampling rate. Movement- and time-derived covariates are also included for all observed and candidate locations.}


\footnotesize
\begin{tabularx}{\linewidth}{p{2.0cm}p{1.2cm}p{1.8cm}p{1.6cm}X}
\toprule
Category & \# Layers & Resolution & Cadence & Examples \\
\midrule

\multirow{2}{*}{\makecell[l]{Weather\\\& climate}}
& \multirow{2}{*}{51} & \multirow{2}{*}{30m -- 28km} & \multirow{2}{*}{1hr -- 8d}\multirow{2}{*}
& \multirow{2}{=}{\makecell[l]{
Temperature, precipitation, wind, snow depth, \\land surface temperature, BioClim~\cite{molteni1996ecmwf,fick2017worldclim,williams2006landsat} 
}} \\
& & & & \\

\midrule

\multirow{2}{*}{\makecell[l]{Vegetation\\\& land cover}}
& \multirow{2}{*}{34} & \multirow{2}{*}{10m -- 500m} & \multirow{2}{*}{3d -- 1yr}
& \multirow{2}{=}{\makecell[l]{
Vegetation cover, vegetation indices, land \\cover, distance to water~\cite{salomonson1989modis,williams2006landsat,brown2022dynamic,pekel2016high}
}} \\
&  &  &  & \\

\midrule

DEM
& 4 & 30m & static & Elevation, slope, aspect, rugosity~\cite{esa2019copernicusdemglo30} \\

\midrule

\multirow{2}{*}{Marine}
& \multirow{2}{*}{6} & \multirow{2}{*}{250m -- 28km} & \multirow{2}{*}{static -- 30d}
& 
\multirow{2}{=}{\makecell[l]{Bathymetry, distance to coast, sea surface temp,\\
chlorophyll, surface current~\cite{chassignet2007hycom,huang2020oisstv21,copernicus2020copernicus} }} \\
& & & & \\

\midrule

Human impact
& 1 & 300m & annual
& WCS Human Footprint Index~\cite{sanderson2022march} \\

\midrule

Embeddings
& 64 & 10m & annual
& \makecell[l]{
AlphaEarth Foundations~\cite{brown2025alphaearth}
} \\

\specialrule{1pt}{1pt}{3pt}

Movement & 3 & -- & -- & Step length, turn angle~\cite{avgar2016integrated} \\

\midrule

Time & 4 & -- & -- & Time of day, day of year~\cite{forrest2025predicting} \\

\bottomrule
\end{tabularx}
\label{tab:movebench_covariates_summary}

\end{table}

\label{sec:covariates}
\textbf{Covariate rasters.} \movebench\ contains 160 publicly-available environmental covariates that are known or hypothesized to influence movement (see \cref{tab:movebench_covariates_summary}). These include features from public environmental and climatic datasets~\cite{molteni1996ecmwf,fick2017worldclim,williams2006landsat,salomonson1989modis} as well as learned geospatial embeddings~\cite{brown2025alphaearth}.
Covariates are downloaded and processed into $101 \times 101$ pixel rasters, centered at observed GPS points and scaled to include appropriate spatial context for a given species and forecast horizon, as in previous raster-based work~\cite{forrest2026predicting} (see Appendix \cref{sec:appendix-covariates}). 
Because we evaluate each method at four different spatiotemporal scales (see \cref{sec:experiments}), we include the full 160-layer raster stack at four different spatial resolutions for each of the 2.6M points in the dataset. In total, the benchmark therefore contains 10.4M raster stacks containing 1.6B individual raster layers representing over two trillion environmental measurements. This is at least four orders of magnitude larger than existing raster-based datasets for animal movement~\cite{forrest2026predicting}, enabling evaluation of predictive methods at a drastically larger scale than was previously possible. 
We also support raster-based versions of common movement- and time-derived covariates such as movement distance and time of movement. See \cref{tab:movebench_covariates_summary} for a summary of all covariates and Appendix \cref{sec:appendix-data} for the full set and acknowledgments.

The use of rasters supports arbitrary definitions of candidate future movement locations (see Appendix \cref{sec:appendix-methods}) as well as emerging raster-based methods~\cite{forrest2026predicting}, but introduces significant engineering complexity. The key challenges involve querying, batching, aligning, and efficiently storing data of different spatial and temporal resolutions from large-scale satellite and climate reanalysis datasets, as well as efficiently loading data during training and evaluation. 
Alongside the processed raster dataset, upon publication we will release the \movebench\ raster download and preparation pipeline as a Python package 
to support future research.

\textbf{Train and test splits.} For each study, we construct one training set and two hold-out test sets representing a \textit{temporal} split and an \textit{individual} split, each capturing a distinct source of ecological variation.
The \textit{temporal} split evaluates generalization to future time by testing on later observations from the same individuals in the training set, reflecting the challenge of predicting movement under changing environmental conditions and possibly different animal life stages.
The \textit{individual} split evaluates generalization across animals by testing on individuals not seen during training, capturing behavioral and environmental heterogeneity among individuals of the same species but within the same population.
We select the held-out time points and individuals in order to mitigate train-test leakage; see Appendix \cref{sec:appendix-traintest}.
In total, \movebench\ contains 1,120,127 training points, 965,233 temporal test points, and 537,781 individual test points.


\textbf{\movebench-Mini.} To enable quicker experimentation, we also make available a ``mini'' set of ten studies across taxonomic groups and movement types. \movebench-Mini includes two migratory birds, two non-migratory birds, two migratory mammals, two non-migratory mammals, one marine species, and one terrestrial reptile. We report study specifics in Appendix \cref{sec:appendix-mini}.

\vspace{-5pt}
\section{Metrics}
\label{sec:metrics}
\vspace{-5pt}

\begin{wrapfigure}{r}{0.38\textwidth}
\vspace{-54pt}
\begin{center}
    \includegraphics[width=0.4\textwidth]{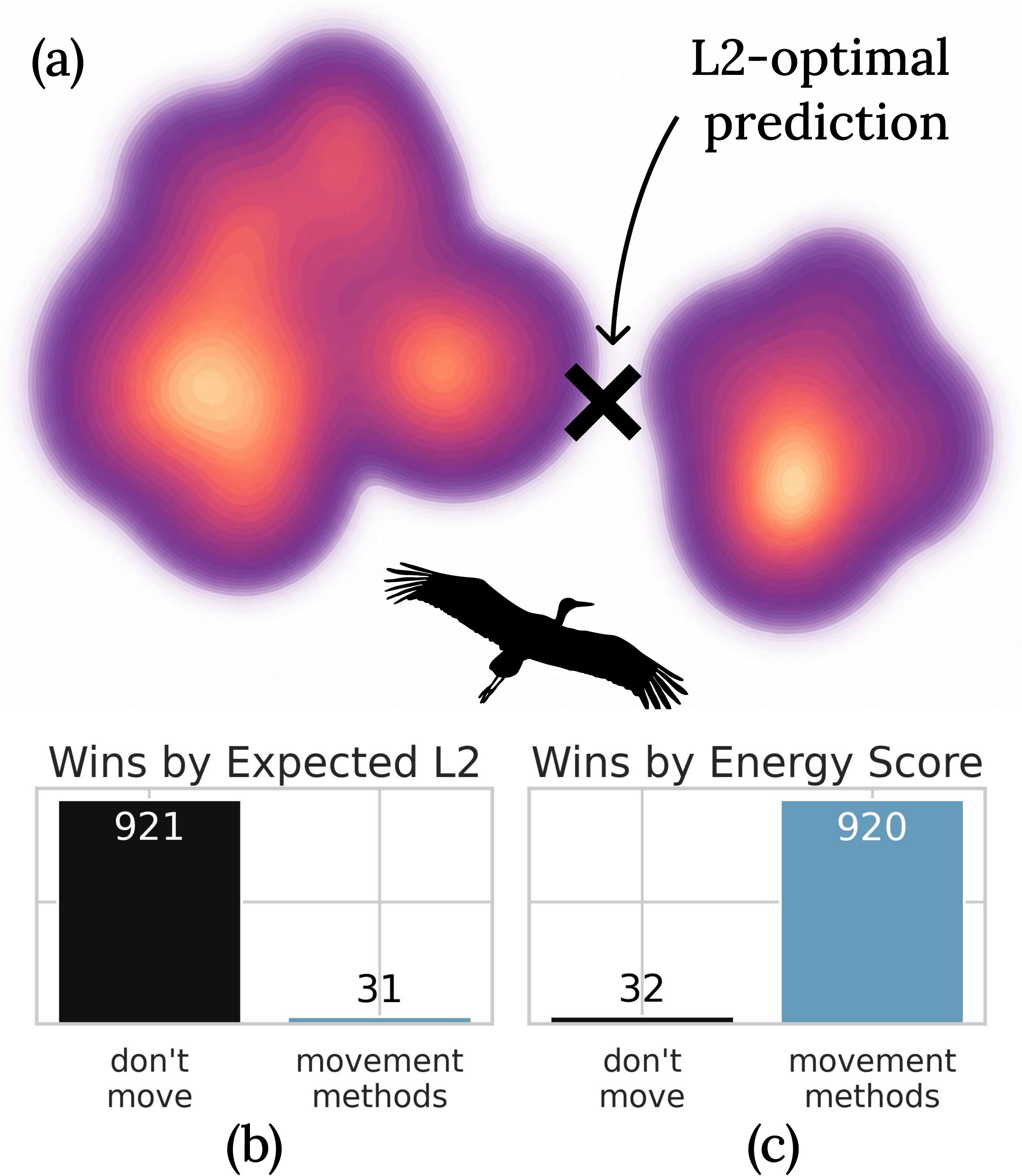}
\end{center}
    \vspace{-10pt}
    \caption{
    \footnotesize 
    \textbf{Probabilistic forecasting metrics.} 
    \textbf{(a)}~Consider an animal with two modes in its next-step movement distribution, \textit{i.e.} it may travel to one of two regions with equal probability. A probabilistic forecast should try to reflect this, but L2-based metrics reward methods that instead deterministically predict the mean or median of the true movement distribution. \textbf{(b)}~We see this empirically on \movebench: according to expected L2 error, the best forecast is to always predict ``don't move''. This is not very useful in practice. \textbf{(c)}~Proper scoring rules like the energy score reward accurate and calibrated probabilistic predictions. In (b) and (c), a ``win'' means that a method has the best score for the given metric on the temporal split for a single study at a particular time horizon.
    }
    \label{fig:metric}
    \vspace{-20pt}
\end{wrapfigure}

Animal movement is inherently \textit{probabilistic}: future motion is often uncertain and multimodal, and thus forecasting methods are designed to return a distribution over possible next steps rather than a single best guess. To evaluate methods, we would ideally compare predicted distributions to the true distribution of the animal’s future movement, of which the observed movement is one realization. This is not possible with commonly used metrics for trajectory forecasting, such as average displacement error~\cite{shi2021sgcn,shi2023tutr,pellegrini2009you}, which evaluate each ground truth point against a single point prediction. 
Further, it is well-known that optimizing for point prediction metrics rewards models that predict a central moment of the true distribution rather than its full shape (\textit{e.g.}, L2 error is minimized by deterministically predicting the geometric median~\cite{dodge1999multivariate}). Thus, when several distinct future destinations are plausible, point prediction metrics can reward an averaged prediction that lies between modes, even if that location itself is implausible~\cite{tyralis2024review,gneiting2007strictly}. See \cref{fig:metric}a.


\textbf{Proper scoring rules} offer a robust framework for assessing the quality of probabilistic forecasts~\cite{waghmare2025proper,gneiting2007probabilistic}. A \textit{scoring rule} is a loss or utility function that can be used to compare predicted distributions with realized observations from a ground truth distribution. A scoring rule is \textit{proper} if it is maximized by predicting the true target distribution rather than hedging toward the median or some otherwise strategically distorted forecast, addressing the problems of point prediction metrics described above (see \cref{fig:metric}b and c). Negative log likelihood (NLL) is one example of a proper scoring rule that is already used widely for comparing wildlife movement forecasts~\cite{forrest2026predicting,cifka2023moveformer}.
However a key limitation of NLL is that it requires access to predicted probability densities, preventing comparisons with deterministic baselines or more complex models where this is expensive or intractable to obtain. 

\textbf{The energy score}~\cite{gneiting2007probabilistic} 
is a proper scoring rule that is particularly well suited to evaluating trajectory forecasts for two reasons: 
(1) It is multivariate, enabling evaluation on \(\mathcal{X}\subseteq \mathbb{R}^2\), and (2) it requires only samples from the predictive distribution, not access to the corresponding predictive density. 
It is a popular choice for evaluating spatial probabilistic forecasts in other applications, such as weather forecasting~\cite{lang2019bivariate,kapoor2023cyclone,mclean2013probabilistic,schuhen2012ensemble}, and has recently been introduced for evaluating probabilistic vehicle trajectory predictions~\cite{shahroudi2024evaluation}.
We adopt it as the main evaluation criterion in \movebench. 

For a given trajectory step $(\ell, t)$ 
and forecast distribution $P_{t,\Delta}$~(see \cref{sec:setup}), the energy score evaluated against \(\ell_{t+\Delta}\), the ground truth location at time $t+\Delta$, is
\begin{equation}
\label{eq:es}
\operatorname{ES}\!\left(P_{t,\Delta},\,\ell_{t+\Delta}\right)
=
\mathbb{E}_{X\sim P_{t,\Delta}}
\bigl\|X-\ell_{t+\Delta}\bigr\|_2
-
\frac{1}{2}
\mathbb{E}_{X,X'\sim P_{t,\Delta}}
\|X-X'\|_2,
\end{equation}
where \(X\) and \(X'\) are i.i.d. draws from the predictive distribution. Lower values are better. For a deterministic predictor, \(P_{t,\Delta}\) is a point mass, so the $\operatorname{ES}$ reduces to the L2 error.

Like many proper scoring rules, the energy score is a combination of two terms. The first can be interpreted as measuring the \textit{accuracy} of a model, and is the average distance between the ground truth location and draws from the predictive distribution (\textit{i.e.}, the expected L2 distance). The second term can be interpreted as measuring the \textit{spread} of a model, computed as half of the average distance between two separate draws from the predictive distribution, and encourages accurate representation of model uncertainty.
A model that is performing well will be both accurate and appropriately calibrated, resulting in a low energy score.
See Appendix \cref{sec:appendix-metrics} for more details on its derivation.

\section{Experiments}
\label{sec:experiments}
\vspace{-5pt}

\textbf{Baselines.} 
We select a set of representative methods to benchmark on \movebench, including both methods that are widely used in practice as well as recent methods incorporating deep learning. The methods are: \textbf{(1)~iSSF Linear}, an integrated step selection function with a standard linear head~\cite{avgar2016integrated}, \textbf{(2)~iSSF MLP}, an iSSF with a multilayer perceptron head, \textbf{(3)~deepSSF}~\cite{forrest2026predicting}, a CNN-based method that operates on spatial rasters, and \textbf{(4)~MoveFormer}~\cite{cifka2023moveformer}, a transformer-based method that incorporates longer temporal context and is trained across studies/species. 
As points of reference, we compare methods with two simple baselines: \textbf{(1)~don't move}, which always predicts that the animal does not move (i.e., \(\hat{\ell}_{t,\Delta} = \ell_t\)); 
and \textbf{(2)~correlated random walk (CRW)}, which samples from the same parametric distributions used to select candidates for SSF-style methods.
We defer detailed descriptions to Appendix \cref{sec:appendix-methods}. 

\begin{figure}[t]
    \centering
    \begin{subfigure}{\linewidth}
        \centering
        \includegraphics[width=\linewidth]{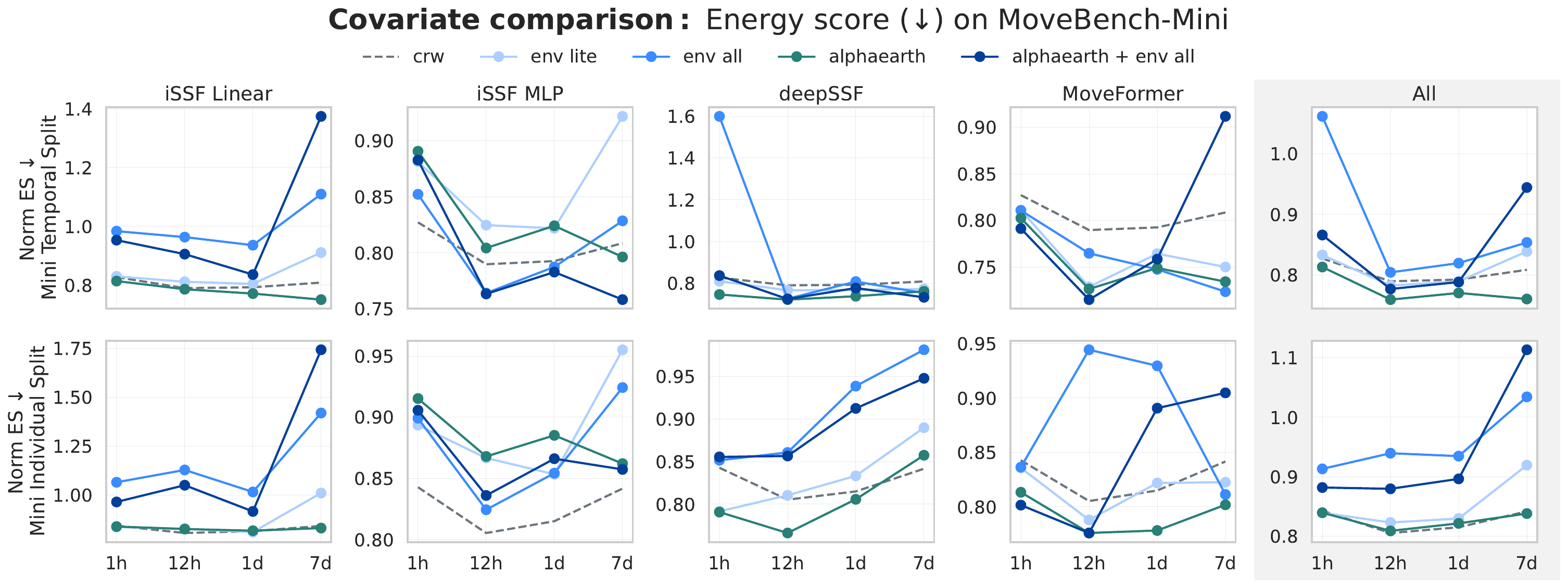}
    \end{subfigure}
    
    \vspace{-4pt}

    \begin{subfigure}{\linewidth}
        \centering
        \includegraphics[width=\linewidth]{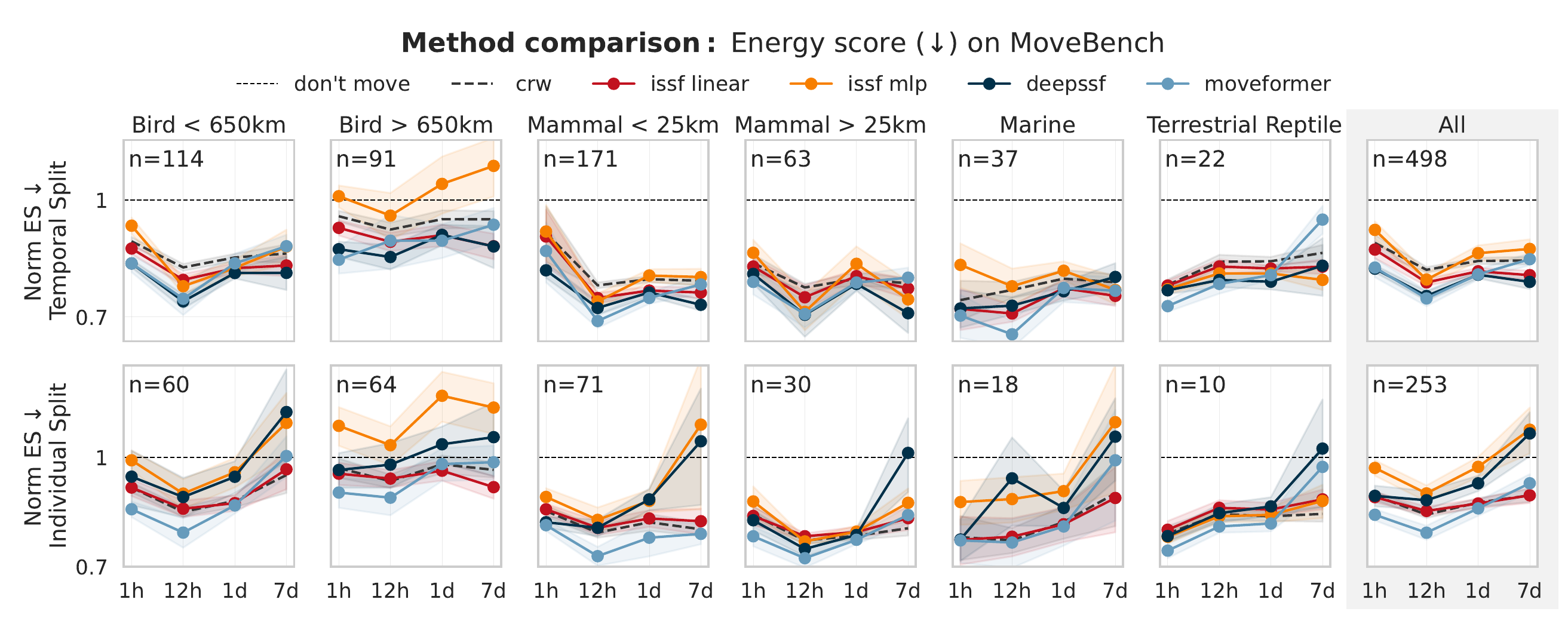}
    \end{subfigure}

    \vspace{-5pt}
    \caption{\small
    \textbf{Main results.} 
    We report energy scores (lower is better, see \cref{sec:metrics}), normalized by average movement distance at the study level to account for inter-study variation and aggregated by method (top) or taxonomic group and travel distance (bottom) (see \cref{sec:experiments}). 
    \textbf{Top (Covariate comparison):} For each model, we compare four different sets of input covariates on \movebench-Mini: ``env lite,'' a small set of common covariates (NDVI and elevation for terrestrial studies, bathymetry and chlorophyll-a for marine studies); ``env all,'' the full set of 96 environmental covariates excluding AlphaEarth; AlphaEarth only; 
    and AlphaEarth + ``env all''. 
    We see that, on average across all methods, AlphaEarth covariates outperform environmental covariates, and that combining AlphaEarth with environmental covariates does not seem to add benefit for most methods.
    These trends are not consistent across methods, however, and in particular iSSF MLP performs best when trained with environmental covariates.
    \textbf{Bottom (Method comparison):} We compare all methods across the entirety of \movebench\ using AlphaEarth covariates (the best-performing set on average), aggregating studies by taxonomic group and median per-individual maximum movement distance. $n$ indicates the number of individuals in each panel. Most methods outperform simple baselines on the temporal split, but not on the individual split, indicating that it is more difficult to generalize to new individuals who were not present in the training set. deepSSF in particular shows a large generalization gap---it is the best-performing overall on the temporal split but worse than the CRW baseline on the individual split. More work is needed to develop methods that perform well across different generalization challenges.
    }
    \label{fig:results}
\end{figure}
\begin{figure}[t]
    \centering
    \includegraphics[width=\linewidth]{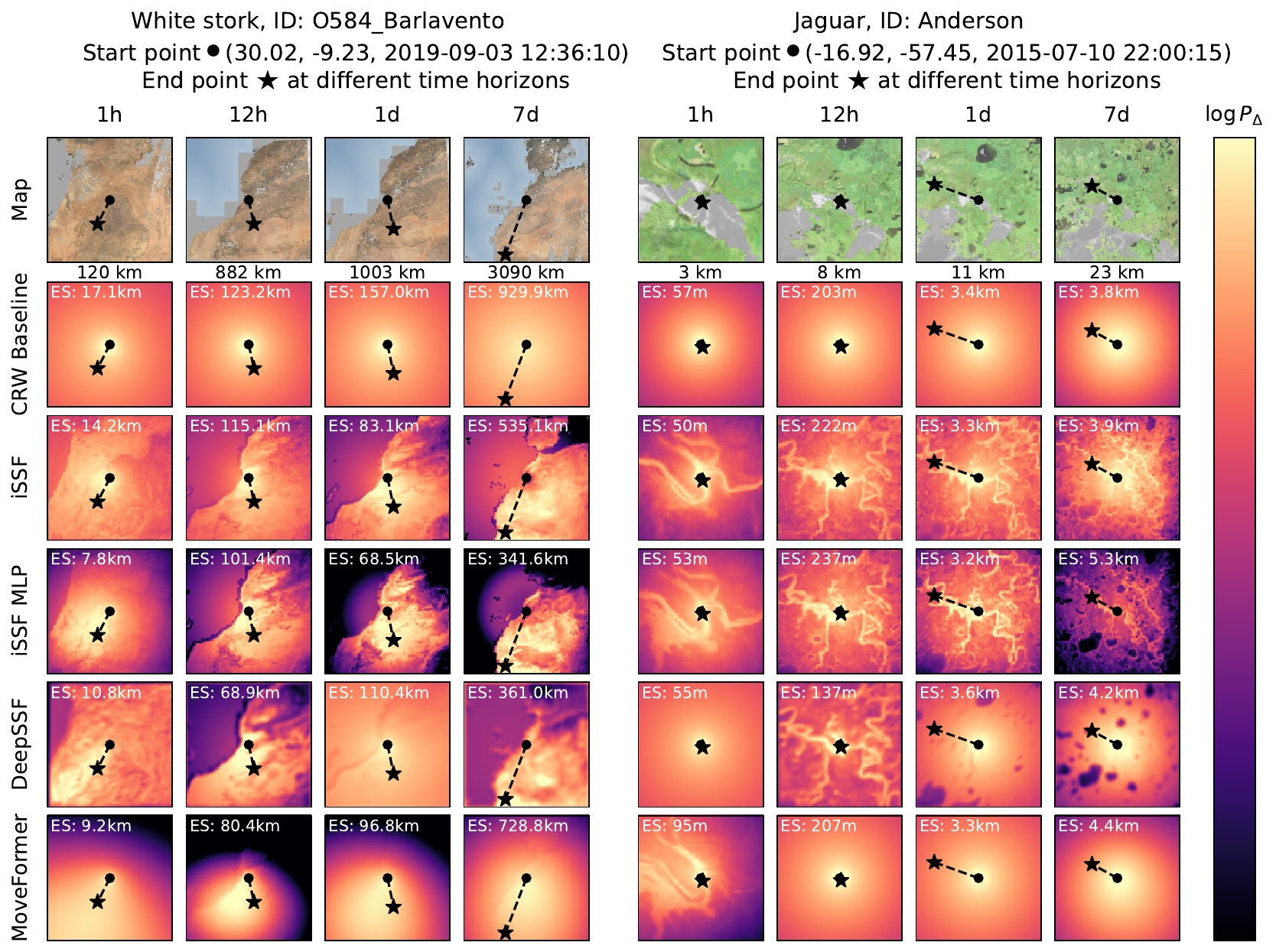}
    \vspace{-18pt}
    \caption{\small
    \textbf{Prediction probability surfaces.} Here we show qualitative examples of forecast distributions of 
    all four methods for all four timescales in our main experiments for two sample points of two species (white stork and jaguar). 
    The black dot represents the animals' starting position and the black star represents their true end position at each time horizon. 
    The map shows Landsat RGB overlaid on Copernicus DEM for the spatial extent available at prediction time---notice that this varies with both study and timescale, and that data from some environmental layers is sometimes missing.
    Prediction heatmaps represent the log predicted probability that the animal moves to any spatial location at the next timestep. 
    Energy scores are displayed for each prediction, computed on the singular sample (lower is better). Notice that some methods' predictions are highly influenced by environmental features (\textit{e.g.,} iSSF, deepSSF), while others focus on broad spatial trends (\textit{e.g.,} MoveFormer).
    }
    \label{fig:qualitative}
\end{figure}

\textbf{Main results.} We train and evaluate each baseline method at a fixed set of forecast horizons $\Delta~\in~\{\text{1 hour}, \text{12 hours}, \text{1 day}, \text{7 days}\}$, compute the energy score (\cref{eq:es}) for each test step at each forecast horizon, and compute a mean energy score for each time horizon for each study. 

We aggregate results across studies by either method (\cref{fig:results} top) or species/range groups (\cref{fig:results} bottom).
All reported results are macro-averages of per-study energy scores. However, because energy score scales with distance, animals that move longer distances on average would dominate a simple average energy score. 
Before aggregating across studies we therefore first normalize each study by the average movement distance of the individuals in that study. We display the mean $\pm$ the standard deviation of these normalized energy scores over each grouping.
Note that normalizing by average movement distance is equivalent to normalizing by the energy score of the ``don't move'' baseline (see \cref{eq:es}), so ``don't move'' always has a normalized ES of 1.
See Appendix \cref{appendix:hparams} for more details on evaluation methodology and complete hyperparameter settings.

\textbf{Covariate comparison.} We first investigate performance of all methods on \movebench-Mini using different sets of environmental covariates in \cref{fig:results} (top).
We compare the full set of covariates available in the benchmark (``alphaearth + env all'') with smaller sets to understand which covariates contribute positively or negatively toward forecasting performance for different models. We see that this is not consistent between models, and that more covariates is not always better, even for deep learning methods. For most methods, as well as on average, AlphaEarth embeddings are the best-performing, while the MLP-based iSSF performs best with a combination of AlphaEarth and environmental covariates. 
While it is not surprising that adding additional highly-correlated covariates to the linear model worsens performance, 
The fact that combining both sources of covariates does not improve performance for other methods---in particular, the deep learning based methods deepSSF and MoveFormer---indicates that more work is needed to enable models to handle large complementary input sets, for instance with additional regularization. 
Rankings of covariates remain fairly stable across the temporal and individual test splits.
Overall, this comparison and the lack of consistency across models point to the importance of covariate selection as a component of methods development.

\textbf{Method comparison.}
We then train and evaluate all methods on all 147 datasets in \movebench\ using AlphaEarth covariates since they were the highest performing on average in the covariate comparison.
In \cref{fig:results} (bottom) we aggregate energy scores over studies based on broad taxonomic group and movement characteristics.
Distance-based groupings are based on the median per-individual maximum displacement of each study. 

We see that no method is consistently best across all taxonomic-range groups or timescales. On in-sample individuals (\textit{i.e.}, the temporal split), most methods usually---but not always---outperform the simple correlated random walk (CRW) baseline, indicating they are leveraging environmental information to improve upon simple movement statistics.
deepSSF---the only method that makes use of full-raster spatial context during training---consistently performs the best on future points of training individuals, though MoveFormer is a very close second-best. All deep learning methods are inconsistent in whether or not they outperform a simple linear iSSF, indicating that deep learning on its own is not enough to guarantee better performance.

We see that trained models generalize better to held-out points in time of individuals in the training set (top row of both plots) compared to held out individuals (bottom row of both plots). 
For held out individuals, most methods do not consistently outperform the correlated random walk baseline. 
Note that the movement parameters of the correlated random walk are also fit to individuals in the training data, so the CRW baseline faces distribution shift as well with held-out individuals. Therefore, methods that outperform CRW in-sample but underperform CRW out-of-sample appear to be leveraging environmental information in a way that is individual-specific and does not generalize well to new individuals. In particular, deepSSF experiences a large performance drop between the temporal and individual splits, making it the best performing for in-sample individuals but the worst performing for out-of-sample individuals.
This indicates that it has learned individual-specific features that are helpful in-distribution and harmful out-of-distribution. MoveFormer---the only method to condition on multiple past timesteps and to train across studies---has the best overall average performance across both hold-out sets.

\textbf{Qualitative analysis.} In \cref{fig:qualitative} we visualize example outputs from an iSSF model on two start points in the dataset to illustrate common qualities of predictive distributions. For each start point, we have a ground truth ``next step'' at each timescale \{1 hour, 12 hours, 1 day, 7 days\}. The plot shows methods' predicted log probability of moving to any available location in the spatial extent. Notice that the spatial extent available to predictive methods increases with timescale according to the movement characteristics of the particular species---for white stork, a migratory bird, its 7 day spatial extent is $26\times$ its 1 hour spatial extent, covering a large swath of Northern African coastline. 
We see that different methods incorporate environmental context as well as movement characteristics differently into their predictions---for instance, MoveFormer shows strong learned directionality in its predictions for both species, and most methods display a preference toward water for the jaguar. 

\vspace{-5pt}
\section{Discussion and Conclusion}
\vspace{-5pt}
\label{sec:discussion}

We introduce \movebench, the first benchmark for global-scale probabilistic animal movement forecasting. The benchmark enables comparisons across the many axes of variation present in animal movement data, including diverse species, spatiotemporal scales, and movement types, and represents a key step towards advancing state-of-the-art in animal movement forecasting. 


\textbf{Research directions \movebench\  highlights as key priorities.}
First, \textit{cross-individual generalization} is a significant challenge across methods. Opportunities include adapting and developing OOD generalization methods, developing mechanisms to better capture individual variation across populations, and further investigating methods to learn movement patterns not just within but also across populations and species. 
Additionally, there is potential to further improve prediction by \textit{incorporating additional biologically meaningful information} into movement prediction, such as life history, individual ID, or inferred behavioral state, as suggested by recent ecology literature~\cite{ranc2022memory,ranc2024role}. 
We also show that \textit{covariate design and selection} is a key component of performance. Opportunities include methods to automatically identify informative covariates per-species, further develop and improve environmental representations from geospatial foundation models with movement forecasting as a target, and address the tradeoffs between interpretability and predictive capacity when using black-box covariates.
Finally, investigating \textit{prediction with long context at any given time horizon} could improve both accuracy and usability of forecasting models, by, \textit{e.g.}, sharing information across spatial and temporal scales via memory (like MoveFormer) or spatial reasoning (like deepSSF).

\textbf{Opportunities to expand movement ecology benchmarking beyond \movebench.} 
We see several exciting future opportunities to build on \movebench, including incorporating larger populations
to understand scaling laws in cross-individual generalization, 
expanding to cover an even broader set of species and regions, 
expanding to include additional target tasks such as behavior segmentation, corridor mapping, or home range estimation, and expanding to include additional modalities often captured concurrently, such as accelerometry.

\textbf{Limitations.} 
\movebench\ is designed for benchmarking probabilistic movement forecasting methods, and should not be used for drawing general \textit{ecological} conclusions without further analysis. Further, data is sourced from existing published movement studies, inheriting any existing taxonomic and geographical sampling biases, and some nuance is inherently lost when averaging across axes of biological variation in the data.
We include further limitations discussion in the Croissant metadata.


\section*{Acknowledgments}

We thank all data owners (see Appendix \cref{sec:appendix-data}), without whom none of this work would be possible. We also thank the organizers and members of MoveBON for support and feedback. 
JK and SaB were supported by NSF award 2330423 and NSERC award 585136. JK was additionally supported by NSF fellowship award 2313998. ShB was supported by a CHE Fellowship for Marine Sciences. JKB was supported by NSF awards 1643532 and NSF 2333609. WR was supported by a Gruber Science Foundation Fellowship. NR was supported by the Harvard University Dean's Fund for Competitive Scientific Research. RK was supported by National Aeronautics and Space Administration Award 80NSSC21K1182. 
RYO was supported by the Kuni Endowed Junior Faculty Fellowship, Hellman Family Faculty Fellowship, and the University of California Early Career Faculty Research Excellence Award.
DT, GM, BK and EA were supported by the Wyoming Center for Wildlife, Technology and Computing (WyldTech).
MO was supported by the Smithsonian Didden Fellowship. CR acknowledges funding from the Gordon and Betty Moore Foundation (GBMF9881; GBMF 11414.01; GBMF12975), and CR and RP from the National Geographic Society (NGS-29784).

\clearpage
\bibliographystyle{plainnat}
\bibliography{bib}

\clearpage
\appendix

\section*{Appendix}
\startcontents[appendix]
\printcontents[appendix]{}{1}{}

\clearpage

\section{Additional related work}

\label{appendix:related}
\textbf{Movement data.} We focus on trajectory data, sequences of location-time tuples that record the movement of some object of interest over time. The most common trajectory datasets in machine learning are GPS ``mobility traces'' of vehicles and pedestrians, typically collected from onboard transmitters~\cite{pkdd-15-predict-taxi-service-trajectory-i,liu2023graphmm,zhu2024unitraj}, or social media check-ins~\cite{yang2016participatory,cho2011friendship,li2024limp,zheng2010geolife}.
Locations are often projected onto a known transportation network/map, reducing the dimensionality of possible movement~\cite{newson2009hidden,wang2022ltp}.

Animal movement data is difficult and costly to collect, clean, and prepare for analysis. Tag deployment often takes place in remote locations, may require veterinary expertise, and 
typically requires government permits and approved ethical protocols.
Significant data cleaning is often required to deal with problems such as irregular or inaccurate data collection due to limited GPS reception~\cite{gupte2022guide,langley2024exmove}. Further, animal trajectory datasets are often analyzed together with environmental covariates such as land cover, elevation, temperature, precipitation, 
or oceanographic variables, which must be derived and aligned from GIS layers, remote sensing products, or weather and climate reanalyses \cite{thurfjell2014applications,northrup2022conceptual}. 

\textbf{Analyzing and modeling movement.}
Advancements in machine-learning-based trajectory analysis have paralleled broader developments in sequence modeling. Early methods drew on statistical state-space and hidden Markov models, later adopting recurrent neural networks and LSTMs for direct sequence prediction \cite{langrock2012flexible,alahi2016social}. More recent approaches use graph neural networks and transformer architectures to improve predictive performance on tasks such as next-step forecasting, multi-step trajectory prediction, and behavior classification
\cite{shi2021sgcn,shi2023tutr,fujii2025towards,wei2024plmtrajrec,zhu2024unitraj}. 

In contrast, recent methods for wildlife movement have modeled the movement process as a function of \textit{habitat selection}, whereby an individual chooses where to go next based in part on the environmental conditions at candidate future locations. Such methods are called \textit{step-selection functions}~\cite{fortin2005wolves,avgar2016integrated}, and are most frequently implemented as a binary classification task where the objective is to differentiate between locations where an animal was actually observed and other nearby locations where it was not, based on environmental conditions.
A range of extensions have been proposed 
for step-selection functions 
to model additional movement processes, including changepoints and behavioral states~\cite{munden2021did,nicosia2017multi,klappstein2023flexible,pohle2024account,bakner2023behavioral}, individual differences or interactions between individuals~\cite{klappstein2024step,muff2020accounting,potts2022assessing, schlagel2019estimating,chatterjee2024modelling}, irregular sampling~\cite{eisaguirre2024rayleigh,hofmann2024methods,munden2021did}, and variation across time and space \cite{valle2024bridging,klappstein2024step,egan2025accounting,forrest2025predicting}. Newer methods have also proposed moving beyond linear models toward deep learning architectures~\cite{forrest2026predicting,cifka2023moveformer,fronville2026analyzing}.
Recently, step-selection functions have been used to predict future movement and space use~\cite{signer2024simulating,rogers2025choices,signer1771estimating,potts2023scale,Hofmann2023connect,cowan2025impact,Westaway2025skinks}, however analysis of their performance remains mostly qualitative and limited to a small number of case studies. Beyond step selection, there have been a small number of papers focused on forecasting movement for single studies, including southern elephant seals~\cite{medina2025improving}, sika deer~\cite{kazama2024sika}, and migratory gulls~\cite{wijeyakulasuriya2020machine}, though none have been assessed more broadly and compared across species. 

\textbf{Evaluating predicted trajectories.}
Trajectory forecasting is commonly evaluated with point-prediction metrics such as the mean L2 distance between predicted points and their associated ground truth observation~\cite{shi2021sgcn,shi2023tutr,pellegrini2009you}. 
However, when future movement is inherently stochastic, point prediction metrics are inappropriate and can reward models that hedge toward the mean or some otherwise strategically distorted forecast~\cite{gneiting2007probabilistic,gneiting2007strictly,smith2015towards,gneiting2011making}. For this reason, probabilistic predictions are commonly evaluated with \textit{proper scoring rules}, which assess the quality of an entire predictive distribution rather than only a single point forecast~\cite{gneiting2007strictly}.
For continuous outcomes, standard examples include the continuous ranked probability score (CRPS) \cite{hersbach2000decomposition}. 

Forecast evaluation in animal movement remains limited, often relying on qualitative inspection~\cite{signer2024simulating} or log loss~\cite{forrest2026predicting,cifka2023moveformer}. While log loss is indeed a proper scoring rule, it requires a likelihood and therefore cannot be used across arbitrary model classes, including deterministic baselines. We therefore utilize the energy score \cite{gneiting2007strictly}, a multivariate sample-based generalization of CRPS, as a probabilistic forecasting metric for wildlife movement data in \movebench.

\textbf{Movement benchmarks.} We provide a summary of existing trajectory benchmarks in \cref{tab:datasets}. Human and vehicle trajectory datasets including GeoLife~\cite{zheng2010geolife}, Porto~\cite{pkdd-15-predict-taxi-service-trajectory-i}, T-Drive~\cite{yuan2010t}, and WorldTrace~\cite{zhu2024unitraj} have large numbers of GPS points, however apart from T-Drive they are quite short, spanning small distances and temporal extents. All human and vehicle trajectory datasets evaluate deterministic map-aligned predictions of future movement. The two existing benchmarks for animal movement data, BEBE~\cite{hoffman2024bebe} and \citet{noonan2019comprehensive}, also do not support probabilistic forecasting. BEBE \cite{hoffman2024bebe} focuses on behavior classification using animal-borne accelerometer tags and does not include GPS data, while \citet{noonan2019comprehensive} anonymizes both GPS locations and timestamps. Neither includes environmental covariates. To our knowledge, \movebench\ is the first benchmark that supports standardized evaluation of probabilistic animal movement forecasting with open-access GPS trajectories and environmental covariates.

\section{Trajectory datasets}
\label{sec:appendix-data}

\subsection{Full list of included studies}

\begingroup
\footnotesize
\setlength{\tabcolsep}{2pt}
\setlength{\LTleft}{0pt}
\setlength{\LTright}{0pt}
\renewcommand{\arraystretch}{1.15}
\begin{longtable}{@{}L{0.05\textwidth} L{0.15\textwidth} L{0.20\textwidth} L{0.17\textwidth} L{0.11\textwidth} L{0.07\textwidth} L{0.10\textwidth} L{0.09\textwidth}@{}}
\caption{MoveBench studies included in the analysis.}\label{tab:movebench_included_studies}\\
\toprule
Index & Study species & Authors & Data & License & N Indiv & GPS fixes & Median fix interval (min) \\
\midrule
\endfirsthead
\caption[]{MoveBench studies included in the analysis (continued).}\\
\toprule
Index & Study species & Authors & Data & License & N Indiv & GPS fixes & Median fix interval (min) \\
\midrule
\endhead
\midrule
\multicolumn{8}{r}{\emph{Continued on next page}}\\
\endfoot
\bottomrule
\endlastfoot
1 & \textit{Mareca penelope} & \citet{van2021MigrationDistanceAffects} & \citet{1054412021StudyEurasianWigeon} & CC0 & 6 & 42,660 & 61 \\
2 & \textit{Anas platyrhynchos} & \citet{kleyheeg2017MovementPatternsOf} & \citet{1054412017MovementPatternsOf} & CC0 & 6 & 3,740 & 60 \\
3 & \textit{Anas platyrhynchos} & \citet{van2013FlexibilityOfContinental} & \citet{1054412013FlexibilityOfContinental} & CC0 & 6 & 3,987 & 360 \\
4 & \textit{Anas platyrhynchos} & \citet{van2018AsTheDuck} & \citet{1054412018AsTheDuck} & CC0 & 6 & 13,785 & 61 \\
5 & \textit{Anser albifrons} & \citet{buitendijk2022MoreGrazingMore} & \citet{1054412023MoreGrazingMore} & CC0 & 6 & 22,337 & 60 \\
6 & \textit{Anser albifrons} & \citet{klzsch2019FlywayConnectivityAnd}\newline \citet{klzsch2020GooseParentsLead} & \citet{1054412019FlywayConnectivityAndc9ae64}\newline \citet{1054412020GooseParentsLead} & CC0 & 6 & 36,919 & 60 \\
7 & \textit{Anser albifrons} & \citet{klzsch2019FlywayConnectivityAnd} & \citet{1054412019FlywayConnectivityAnd} & CC0 & 6 & 13,297 & 60 \\
8 & \textit{Anser anser} & \citet{buitendijk2022MoreGrazingMore} & \citet{1054412023MoreGrazingMore} & CC0 & 6 & 53,183 & 60 \\
\end{longtable}
\endgroup

\subsection{Trajectory curation and filtering details}

We start by aggregating movement \textit{studies}---single datasets that have been previously released by movement researchers (\textit{i.e.}, a row in \cref{tab:movebench_included_studies}). A single study contains GPS trajectory data for one or more individuals of one or more species. Because different species exhibit different movement patterns, they are often tracked using different hardware with different settings. If a single dataset contained multiple species, we split the dataset and consider each species a separate ``study'' for all data aggregation and analysis, and refer to these as so in the main text.

We begin with the 215 studies containing GPS data on the Movebank Data Repository~\cite{movebankMovebank}, a subset of data from Movebank~\cite{web205Movebank} that have Creative Commons licensing, DOIs, and have already undergone some quality assurance and anomaly filtering. We then add an additional 14 studies sourced from a literature review of recent movement methods papers that performed some kind of empirical study, as well as 5 additional suggested studies from domain experts in the Movement Biodiversity Observation Network (Move BON)\cite{geobonMove} Working Group on AI. In total this results in 234 studies and 89M raw GPS points that we filter down to the final set as follows.

To make this more tractable for benchmarking, we apply a set of coarse filtering criteria followed by manual inspection of all points. We log all dataset operations for reproducibility and report the full set of modifications in the dataset Croissant file (see \cref{sec:appendix-data-graph}).

We first thin the GPS points to a maximum of one point per hour, matching the minimum forecast horizon we evaluate (see \cref{sec:metrics}), leaving us with 14M points. We then filter out individuals with $< 200$ GPS points, and keep a maximum of $17,520$ GPS points per individual. We chose the latter number as the number of hours in two years, allowing for a temporal train/test split to contain a full annual cycle of behavioral patterns in both train and test (see ``Train and test splits'' below). If an individual has more than the maximum number of points, we keep the most recent $17,520$. Many datasets did not meet these criteria; our final set for the benchmark includes 147 studies.

We then inspect the remaining data by hand with the goal of selecting six individuals per species per study. 
To do this, we use a custom web application we developed, which visualizes trajectories spatially (on a map) and temporally (on a timeline). We inspect for remaining anomalous data, such as trajectories that are stationary for extended periods of time due to falling off the animal, and remove affected individuals. 
We then keep all individuals whose temporal coverage allows for a temporal train/test split (see below). Finally, if there are more than six individuals remaining, we randomly sample six to keep, discarding the rest.

Note that these decisions were made specifically to keep the dataset of a tractable size. In the future, users may choose to train on larger, less filtered raw datasets.

\subsection{Additional details on train/test split}
\label{sec:appendix-traintest}

\textbf{Train and test splits.} For each study pair, we aim to create one training set and two hold-out test sets representing a \textit{temporal} split and an \textit{individual} split. 

The temporal split defines a hold-out test set that includes test data that comes from future points in time with respect to the training data. We define a single split timestamp for each individual---any observations before that time goes in the training set, and any observations after go in the temporal test set. We aim for this split to be roughly 50/50. When individuals have overlapping spatial and temporal ranges, we avoid train-test leakage by enforcing that every individual in the study population has the same split time, so that models cannot leverage memorized information from nearby individuals during test time. For some studies, observations did not cover a large enough temporal range that would allow movement patterns to appear in both the training and temporal test set---for instance, splitting the data from long-distance migrants with only one recorded migration would mean that training data may come from completely disjoint countries (or even continents) than temporal test data. For these individuals, we only include an individual test set, and do not include a temporal test set.

The individual split defines a hold-out test set containing individuals that do not appear in the training set. We select one-third of the individuals from each study to be held-out, rounding down if there are an odd number of individuals. To avoid train-test leakage, for hold-out individuals we test on data from the same time range that would otherwise be in their temporal hold-out set---\textit{i.e.}, any data for hold-out individuals that occurs before the temporal split point remains unused during both training and testing.

To avoid two training sets per species-study pair, the final training set is the intersection of the training data that would result from either split being created on its own---that is, it includes all data before the temporal split point for all individuals not in the held-out individuals set.

\subsection{Data modification records}
\label{sec:appendix-data-graph}

We built a custom application for inspecting and processing data from its raw state into the version in \movebench. The application logs atomic records of every data modification operation to maintain full data provenance. We include the complete set of data modification records alongside the release in the Croissant file.

\subsection{\movebench-Mini}
\label{sec:appendix-mini}

The studies in \movebench-Mini are: \citet{forrest2025predicting}, \citet{borchering2017ResourceDrivenEncounters}, \citet{garthe2022EffectsOfAgricultural},
  \citet{aikens2024ChallengingConventionalViews}, \citet{costa2024TwoDecadesOf}, \citet{accio2022TimingIsCritical}, \citet{bartlambrooks2013InSearchOf},
  \citet{gatt2019PreLayingMovements}, \citet{dodge2014EnvironmentalDriversOf}, and \citet{bastillerousseau2019MigrationTriggersIn}.

\section{Environmental covariates}
\label{sec:appendix-covariates}

\subsection{Full list of included covariates}


\begingroup
\scriptsize
\setlength{\tabcolsep}{1.8pt}
\setlength{\LTleft}{0pt}
\setlength{\LTright}{0pt}
\renewcommand{\arraystretch}{1.08}

\paragraph*{Weather \& climate}
\begin{longtable}{@{}L{0.18\textwidth} L{0.25\textwidth} L{0.07\textwidth} L{0.08\textwidth} L{0.14\textwidth} L{0.22\textwidth}@{}}
\caption{MoveBench environmental covariates in the Weather \& climate group, with data source, resolution, temporal coverage, and license. ERA5 entries cite \citet{hersbach2018era5} and were downloaded from \citet{copernicus2026era5}. The results contain modified Copernicus Climate Change Service information 2020. Neither the European Commission nor ECMWF is responsible for any use that may be made of the Copernicus information or data it contains.}\label{tab:movebench_covariates_weather_and_climate}\\
\toprule
Source & Covariate & Spatial & Temporal & Coverage & License \\
\midrule
\endfirsthead
\caption[]{MoveBench environmental covariates in the Weather \& climate group (continued).}\\
\toprule
Source & Covariate & Spatial & Temporal & Coverage & License \\
\midrule
\endhead
\midrule
\multicolumn{6}{r}{\emph{Continued on next page}}\\
\endfoot
\bottomrule
\endlastfoot
ERA5-Land {\footnotesize (\citet{era5Land})} & Hourly Temperature 2 m & 11km & 1 hour & 1950-2026 & CC-BY-4.0 \\
 & Daily Temperature 2 m & 11km & 1 day & 1950-2026 &  \\
 & Hourly Total Precipitation & 11km & 1 hour & 1950-2026 &  \\
 & Daily Total Precipitation & 11km & 1 day & 1950-2026 &  \\
 & Hourly Dewpoint Temperature 2 m & 11km & 1 hour & 1950-2026 &  \\
 & Daily Dewpoint Temperature 2 m & 11km & 1 day & 1950-2026 &  \\
 & Hourly Surface Pressure & 11km & 1 hour & 1950-2026 &  \\
 & Daily Surface Pressure & 11km & 1 day & 1950-2026 &  \\
 & Hourly U Component of Wind 10 m & 11km & 1 hour & 1950-2026 &  \\
 & Daily U Component of Wind 10 m & 11km & 1 day & 1950-2026 &  \\
 & Hourly V Component of Wind 10 m & 11km & 1 hour & 1950-2026 &  \\
 & Daily V Component of Wind 10 m & 11km & 1 day & 1950-2026 &  \\
 & Hourly Surface Solar Radiation Downwards & 11km & 1 hour & 1950-2026 &  \\
 & Daily Surface Solar Radiation Downwards & 11km & 1 day & 1950-2026 &  \\
 & Hourly Surface Thermal Radiation Downwards & 11km & 1 hour & 1950-2026 &  \\
 & Daily Surface Thermal Radiation Downwards & 11km & 1 day & 1950-2026 &  \\
 & Hourly Surface Net Solar Radiation & 11km & 1 hour & 1950-2026 &  \\
 & Daily Surface Net Solar Radiation & 11km & 1 day & 1950-2026 &  \\
 & Hourly Surface Net Thermal Radiation & 11km & 1 hour & 1950-2026 &  \\
 & Daily Surface Net Thermal Radiation & 11km & 1 day & 1950-2026 &  \\
 & Hourly Surface Latent Heat Flux & 11km & 1 hour & 1950-2026 &  \\
 & Daily Surface Latent Heat Flux & 11km & 1 day & 1950-2026 &  \\
 & Hourly Surface Sensible Heat Flux & 11km & 1 hour & 1950-2026 &  \\
 & Daily Surface Sensible Heat Flux & 11km & 1 day & 1950-2026 &  \\
 & Hourly Snow Depth & 11km & 1 hour & 1950-2026 &  \\
 & Daily Snow Depth & 11km & 1 day & 1950-2026 &  \\
ERA5 {\footnotesize (\citet{hersbach2018era5}; downloaded from \citet{copernicus2026era5})} & Boundary Layer Height & 28km & 1 hour & 1950-2026 & CC-BY-4.0 \\
 & Total Cloud Cover & 28km & 1 hour & 1950-2026 &  \\
 & Mean Sea Level Pressure & 28km & 1 hour & 1950-2026 &  \\
 & Thermal uplift & 11km & Hourly & 1950-2026 (derived) &  \\
Landsat {\footnotesize (\citet{usgsLandsat})} & Land surface temperature & 30m & 8 day & 1984-2026 & public domain \\
Bioclim {\footnotesize (\citet{bioclim})} & BIO1: Annual Mean Temperature & 1km & Static & 1970-2000 climatology; static layer & CC-BY-4.0 \\
 & BIO2: Mean Diurnal Range & 1km & Static & 1970-2000 climatology; static layer &  \\
 & BIO3: Isothermality & 1km & Static & 1970-2000 climatology; static layer &  \\
 & BIO4: Temperature Seasonality & 1km & Static & 1970-2000 climatology; static layer &  \\
 & BIO5: Max Temperature of Warmest Month & 1km & Static & 1970-2000 climatology; static layer &  \\
 & BIO6: Min Temperature of Coldest Month & 1km & Static & 1970-2000 climatology; static layer &  \\
 & BIO7: Temperature Annual Range & 1km & Static & 1970-2000 climatology; static layer &  \\
 & BIO8: Mean Temperature of Wettest Quarter & 1km & Static & 1970-2000 climatology; static layer &  \\
 & BIO9: Mean Temperature of Driest Quarter & 1km & Static & 1970-2000 climatology; static layer &  \\
 & BIO10: Mean Temperature of Warmest Quarter & 1km & Static & 1970-2000 climatology; static layer &  \\
 & BIO11: Mean Temperature of Coldest Quarter & 1km & Static & 1970-2000 climatology; static layer &  \\
 & BIO12: Annual Precipitation & 1km & Static & 1970-2000 climatology; static layer &  \\
 & BIO13: Precipitation of Wettest Month & 1km & Static & 1970-2000 climatology; static layer &  \\
 & BIO14: Precipitation of Driest Month & 1km & Static & 1970-2000 climatology; static layer &  \\
 & BIO15: Precipitation Seasonality & 1km & Static & 1970-2000 climatology; static layer &  \\
 & BIO16: Precipitation of Wettest Quarter & 1km & Static & 1970-2000 climatology; static layer &  \\
 & BIO17: Precipitation of Driest Quarter & 1km & Static & 1970-2000 climatology; static layer &  \\
 & BIO18: Precipitation of Warmest Quarter & 1km & Static & 1970-2000 climatology; static layer &  \\
 & BIO19: Precipitation of Coldest Quarter & 1km & Static & 1970-2000 climatology; static layer &  \\
Copernicus DEM GLO-30 {\footnotesize (\citet{copernicusDEM})} + ERA5-Land {\footnotesize (\citet{era5Land})} & Orographic uplift & 30m & Hourly & 1950-2026 (derived from ERA5-Land) & Copernicus DEM GLO-30 \& Free and open Copernicus WorldDEM-30 licence; attribution required.; CC-BY-4.0 \\
\end{longtable}

\paragraph*{Vegetation \& land cover}
\begin{longtable}{@{}L{0.18\textwidth} L{0.25\textwidth} L{0.07\textwidth} L{0.08\textwidth} L{0.14\textwidth} L{0.22\textwidth}@{}}
\caption{MoveBench environmental covariates in the Vegetation \& land cover group, with data source, resolution, temporal coverage, and license.}\label{tab:movebench_covariates_vegetation_and_land_cover}\\
\toprule
Source & Covariate & Spatial & Temporal & Coverage & License \\
\midrule
\endfirsthead
\caption[]{MoveBench environmental covariates in the Vegetation \& land cover group (continued).}\\
\toprule
Source & Covariate & Spatial & Temporal & Coverage & License \\
\midrule
\endhead
\midrule
\multicolumn{6}{r}{\emph{Continued on next page}}\\
\endfoot
\bottomrule
\endlastfoot
Landsat {\footnotesize (\citet{usgsLandsat})} & NDVI & 30m & 8 day & 1984-2026 & public domain \\
 & EVI & 30m & 8 day & 1984-2026 &  \\
 & NDSI (snow) & 30m & 8 day & 1984-2026 &  \\
 & BAI (burn) & 30m & 8 day & 1984-2026 &  \\
MODIS MOD44B.061 {\footnotesize (\citet{modisMod44bV061})} & Percent Tree Cover & 250m & 1 year & 2000-2024 & CC0 \\
 & Percent Non-tree Vegetation & 250m & 1 year & 2000-2024 &  \\
 & Percent Non-vegetated & 250m & 1 year & 2000-2024 &  \\
MODIS MCD12Q1.061 LC\_Type1 {\footnotesize (\citet{modisMcd12q1V061})} & LC Type1 01: Evergreen Needleleaf Forests & 500m & 1 year & 2001-2024 & CC0 \\
 & LC Type1 02: Evergreen Broadleaf Forests & 500m & 1 year & 2001-2024 &  \\
 & LC Type1 03: Deciduous Needleleaf Forests & 500m & 1 year & 2001-2024 &  \\
 & LC Type1 04: Deciduous Broadleaf Forests & 500m & 1 year & 2001-2024 &  \\
 & LC Type1 05: Mixed Forests & 500m & 1 year & 2001-2024 &  \\
 & LC Type1 06: Closed Shrublands & 500m & 1 year & 2001-2024 &  \\
 & LC Type1 07: Open Shrublands & 500m & 1 year & 2001-2024 &  \\
 & LC Type1 08: Woody Savannas & 500m & 1 year & 2001-2024 &  \\
 & LC Type1 09: Savannas & 500m & 1 year & 2001-2024 &  \\
 & LC Type1 10: Grasslands & 500m & 1 year & 2001-2024 &  \\
 & LC Type1 11: Permanent Wetlands & 500m & 1 year & 2001-2024 &  \\
 & LC Type1 12: Cropland. & 500m & 1 year & 2001-2024 &  \\
 & LC Type1 13: Urban and Built-up Lands & 500m & 1 year & 2001-2024 &  \\
 & LC Type1 14: Cropland/Natural Vegetation Mosaics & 500m & 1 year & 2001-2024 &  \\
 & LC Type1 15: Permanent Snow and Ice & 500m & 1 year & 2001-2024 &  \\
 & LC Type1 16: (sand, rock, soil) areas with less than 10\% vegetation. & 500m & 1 year & 2001-2024 &  \\
 & LC Type1 17: Water Bodies & 500m & 1 year & 2001-2024 &  \\
Dynamic World {\footnotesize (\citet{dynamicworld})} & Water & 10m & 3 to 5 days & 2015-2026 & CC-BY-4.0 \\
 & Trees & 10m & 3 to 5 days & 2015-2026 &  \\
 & Grass & 10m & 3 to 5 days & 2015-2026 &  \\
 & Flooded Vegetation & 10m & 3 to 5 days & 2015-2026 &  \\
 & Crops & 10m & 3 to 5 days & 2015-2026 &  \\
 & Shrub and Scrub & 10m & 3 to 5 days & 2015-2026 &  \\
 & Built & 10m & 3 to 5 days & 2015-2026 &  \\
 & Bare & 10m & 3 to 5 days & 2015-2026 &  \\
 & Snow and Ice & 10m & 3 to 5 days & 2015-2026 &  \\
JRC Global Surface Water {\footnotesize (\citet{jrcGlobalwater})} & Distance to water & 30m & Mostly static & 1984-2021 source observations; derived layer used as static & JRC Global Surface Water \& Free and open use under Copernicus / European Commission reuse terms; attribution required. \\
\end{longtable}

\paragraph*{DEM}
\begin{longtable}{@{}L{0.18\textwidth} L{0.25\textwidth} L{0.07\textwidth} L{0.08\textwidth} L{0.14\textwidth} L{0.22\textwidth}@{}}
\caption{MoveBench environmental covariates in the DEM group, with data source, resolution, temporal coverage, and license.}\label{tab:movebench_covariates_dem}\\
\toprule
Source & Covariate & Spatial & Temporal & Coverage & License \\
\midrule
\endfirsthead
\caption[]{MoveBench environmental covariates in the DEM group (continued).}\\
\toprule
Source & Covariate & Spatial & Temporal & Coverage & License \\
\midrule
\endhead
\midrule
\multicolumn{6}{r}{\emph{Continued on next page}}\\
\endfoot
\bottomrule
\endlastfoot
Copernicus DEM GLO-30 {\footnotesize (\citet{copernicusDEM})} & Elevation & 30m & Static & 2010-2015 acquisition window; static layer & Copernicus DEM GLO-30 \& Free and open Copernicus WorldDEM-30 licence; attribution required. \\
 & Slope & 30m & Static & 2010-2015 acquisition window; static layer &  \\
 & Aspect & 30m & Static & 2010-2015 acquisition window; static layer &  \\
 & Rugosity/ruggedness & 30m & Static & 2010-2015 acquisition window; static layer &  \\
\end{longtable}

\paragraph*{Human impact}
\begin{longtable}{@{}L{0.18\textwidth} L{0.25\textwidth} L{0.07\textwidth} L{0.08\textwidth} L{0.14\textwidth} L{0.22\textwidth}@{}}
\caption{MoveBench environmental covariates in the Human impact group, with data source, resolution, temporal coverage, and license.}\label{tab:movebench_covariates_human_impact}\\
\toprule
Source & Covariate & Spatial & Temporal & Coverage & License \\
\midrule
\endfirsthead
\caption[]{MoveBench environmental covariates in the Human impact group (continued).}\\
\toprule
Source & Covariate & Spatial & Temporal & Coverage & License \\
\midrule
\endhead
\midrule
\multicolumn{6}{r}{\emph{Continued on next page}}\\
\endfoot
\bottomrule
\endlastfoot
WCS {\footnotesize (\citet{wcsHumanFootprint}; \citet{wcsHumanFootprint2020})} & Human impact index & 300m & Annual & 2001-2020 & CC BY-NC-SA 3.0 \\
\end{longtable}

\paragraph*{Marine}
\begin{longtable}{@{}L{0.18\textwidth} L{0.25\textwidth} L{0.07\textwidth} L{0.08\textwidth} L{0.14\textwidth} L{0.22\textwidth}@{}}
\caption{MoveBench environmental covariates in the Marine group, with data source, resolution, temporal coverage, and license.}\label{tab:movebench_covariates_marine}\\
\toprule
Source & Covariate & Spatial & Temporal & Coverage & License \\
\midrule
\endfirsthead
\caption[]{MoveBench environmental covariates in the Marine group (continued).}\\
\toprule
Source & Covariate & Spatial & Temporal & Coverage & License \\
\midrule
\endhead
\midrule
\multicolumn{6}{r}{\emph{Continued on next page}}\\
\endfoot
\bottomrule
\endlastfoot
Copernicus Global Ocean Waves {\footnotesize (\citet{copernicusGlobalOcean})} & Bathymetry (sea floor depth) & 0.9km & Static & 2022-2026 & Copernicus Marine Service \& Free and open use under Copernicus Marine Service licence; attribution required. \\
MOD44W.006 {\footnotesize (\citet{carroll2017mod44w})} & Distance to coast & 250m & Static & 2000-2015 source observations; derived layer used as static & CC0 \\
NOAA CDR {\footnotesize (\citet{noaaCDR})} & Sea surface temperature & 28km & 30 day & 1981-2026 & NOAA OISST v2.1 \& Free NOAA/NCEI data; citation requested; no warranty \\
Copernicus Satellite Ocean Color {\footnotesize (\citet{copernicusGlobalOceanColour})} & Ocean surface chlorophyll & 4000m & 30 day & 1997-2026 & Global Ocean Colour \& Free Copernicus Marine Service licence; redistribution and derivative products permitted. \\
HYCOM {\footnotesize (\citet{hycom})} & Surface current eastward & 8900m & 1 day & 1992-2024 & Freely available with no restrictions \\
 & Surface current northward & 8900m & 1 day & 1992-2024 &  \\
\end{longtable}

\paragraph*{Embeddings}
\begin{longtable}{@{}L{0.18\textwidth} L{0.25\textwidth} L{0.07\textwidth} L{0.08\textwidth} L{0.14\textwidth} L{0.22\textwidth}@{}}
\caption{MoveBench environmental covariates in the Embeddings group, with data source, resolution, temporal coverage, and license.}\label{tab:movebench_covariates_embeddings}\\
\toprule
Source & Covariate & Spatial & Temporal & Coverage & License \\
\midrule
\endfirsthead
\caption[]{MoveBench environmental covariates in the Embeddings group (continued).}\\
\toprule
Source & Covariate & Spatial & Temporal & Coverage & License \\
\midrule
\endhead
\midrule
\multicolumn{6}{r}{\emph{Continued on next page}}\\
\endfoot
\bottomrule
\endlastfoot
AlphaEarth {\footnotesize (\citet{brown2025alphaearth})} & AlphaEarth - 64 bands & 10m & Annual & 2017-2025 & CC-BY 4.0 \\
\end{longtable}
\endgroup

\subsection{Details on raster preparation}

Raster preparation begins by defining a \emph{raster recipe} for a given benchmark setting. This recipe defines the spatial characteristics of the rasters that will be downloaded for each study. In particular, it specifies four things: which environmental variables we need; which of them are time-varying versus static; how much spatial context we retain around each movement location; and what spatial resolution is practical for storing and reusing the resulting rasters. 
We define this recipe before any download occurs so that all later stages, including feature extraction and raster-based models, operate on a consistent set of bands, extents, and spatial grids. We can therefore reuse the same raw raster cache across multiple downstream experiments whenever they share the same spatial recipe, even if they use different models or different subsets of
the bands.

We resolve the recipe separately for each benchmark timescale. We determine the scale (meters per pixel) and extent (meters per raster) based on each study's movement statistics at that timescale. We elaborate on this more below.
We do this because the relevant movement radius grows with forecast horizon, and storing rasters at their native resolution for very large extents would not be tractable.

\paragraph{Raster sources and layer families.}
Most layers are downloaded from Google Earth Engine. The layers that are not available on Google Earth Engine---Human Footprint Index and BioClim---are downloaded in full as local GeoTIFF archives.

\paragraph{Timescale-specific extent and scale.}
We determine both raster extent and raster scale from movement statistics in the training split at the target forecast horizon. We first identify valid interval steps after partitioning each
trajectory into temporally coherent ``bursts'' (segments of uninterrupted movement). By default, we treat inter-fix gaps above the 0.999 quantile of an individual's revisit-time distribution as burst boundaries, so very large
temporal gaps do not create artificial long-distance steps. We then compute a conservative spatial radius $b$ from the valid training-step lengths \(L\):
\[
b = \min\!\left(2.0 \times Q_{0.95}(L),\; Q_{0.99}(L)\right),
\]
where \(Q_q(L)\) denotes the \(q\)-quantile. This is the spatial range we consider as ``available'' to the animal during its next step, and is equal to the minimum of double its 95th percentile step length and its 99th percentile step length. 
This ensures that nearly all observed steps will have raster coverage while reducing the influence of any outliers. We use \(b\) as the half-width of the
square spatial context around a focal point, so each local raster patch covers a square of side length \(2b\).

We choose the stored resolution for each layer by anchoring it to a canonical downstream chip size of \(101\times 101\) pixels to match prior work~\cite{forrest2026predicting}. 
If a chip has half-width \(b\), then a \(101\times 101\) grid
spans the full width \(2b\) with 101 pixels across each side, so the corresponding pixel size is \(2b/101\). 
For a layer with native resolution \(s_{\mathrm{native}}\), we
therefore store it at
\[
s=\mathrm{clip}\!\left(s_{\mathrm{native}},\; \frac{2b}{101},\; \frac{2b}{3}\right).
\]
This rule makes the raw cache match the scale of the canonical chip whenever the native raster is finer than necessary, while still enforcing a minimum of about \(3\times 3\) pixels of
spatial support for very coarse layers. The lower clamp, \(2b/101\), prevents fine-resolution products such as AlphaEarth (10m resolution) or DEM (30m resolution) from producing source chips that are much denser
than the eventual \(101\times 101\) grid at long horizons, which would waste storage. The upper clamp, \(2b/3\), prevents coarse products such as ERA5 fields (11--28km resolution) from collapsing to one or two pixels across the chip at short horizons. We apply this rule independently to each layer because their native resolutions differ by orders of magnitude.

We use two storage modes, depending on how a layer varies in time and how much reuse we expect across points: (1) We store \emph{chips} for layers whose value depends on the focal timestamp and therefore must be queried separately for each point-time pair. In our benchmark, this includes layers with high temporal resolution such as ERA5-Land hourly and daily fields, ERA5 hourly fields, Landsat composites and their derived indices, Dynamic World probabilities, and time-varying marine layers. Each chip is a point-centered square crop with half-width \(b\). (2) We store study-level \textit{tiles}, reusable larger rasters for layers that are static or annual and can therefore serve many points within the same study. In our benchmark, this includes DEM, distance-to-water, bathymetry, distance-to-shore, annual MODIS land cover, annual MODIS vegetation fractions, and
annual AlphaEarth embeddings. 


\paragraph{Earth Engine query rules.}
We query different layers with source-specific temporal rules.
For ERA5-Land hourly layers, we select the nearest image within \(\pm 1\) hour of the focal timestamp. If ERA5-Land leaves any pixels masked, we fill those masked pixels from the
corresponding bands in the ERA5 hourly collection. For ERA5-Land daily aggregates, we select the nearest image within \(\pm 1\) day. If the daily aggregate is masked, we fill the six
meteorological bands that also exist in the ERA5 daily collection---2 m air temperature, 2 m dewpoint temperature, total precipitation, surface pressure, and the two 10 m wind
components---from ERA5 daily. The radiative flux bands and snow depth remain missing when no ERA5-Land daily aggregate is available.

For layers that are composites over some time range $T$ (\textit{e.g.} for Landsat, $T =$ 8-days), we query the most recent composite in the interval \([t-T\text{}, t]\).  For Landsat, if the 8-day composite is missing or
masked at some pixels, we fill the missing pixels from the most recent 32-day composite in \([t-32\text{ days}, t]\). We then derive NDVI, EVI, NDWI, NDSI, and BAI from the multispectral bands after loading.

For NOAA OISST, we compute a 30-day mean and a 30-day standard deviation of sea-surface temperature over
the interval \([t-30\text{ days}, t]\). For Copernicus chlorophyll, we compute a 30-day mean over the same interval. For HYCOM currents, we select the latest image in \([t-1\text{ day},
t]\) and derive current speed as \(\sqrt{u^2+v^2}\) from the eastward and northward surface components.

Dynamic World requires special handling because the product begins on 2015-06-23 and because cloud-free daily coverage is irregular. If the focal timestamp is earlier than 2015-06-23, we
use the median Dynamic World probability image over the first year of product availability, 2015-06-23 through 2016-06-23. Otherwise, we compute a median over a \(\pm 15\)-day window around
the focal timestamp, and if that window contains no imagery we widen it to \(\pm 30\) days.

For annual shared layers, we resolve the year from the focal timestamp and then snap to the nearest available year in the source archive when the exact year is missing. This rule applies to
annual MODIS land cover, annual MODIS vegetation fractions, and annual AlphaEarth embeddings. Within the resolved year, we use a earliest available annual image.

For shared static layers, we query fixed products directly. We obtain elevation by mosaicking the Copernicus GLO-30 DEM. We derive distance to water from the JRC Global Surface Water
\texttt{max\_extent} mask using cumulative geodesic cost and clip the result at 30 km. We derive distance to shore from the latest MOD44W \texttt{water\_mask}, computing cumulative cost
from land and clipping the result at 100 km. We read bathymetric depth from the static Copernicus marine layer.

\paragraph{Planning and batching shared tiles.}
For \textit{tiles} that are shared between many training/evaluation points, we do not center a shared raster tile on every point. Instead, we first group points into coarse longitude-latitude buckets, separately within each data split. We then expand each bucket by
the movement radius \(b\), handle antimeridian crossing explicitly when needed, and convert the resulting buffered rectangles into a regular global tile grid whose cell size is \(350 \times
s\) square meters. We keep only tiles that intersect at least one buffered rectangle. For annual layers, we reuse the same spatial tile plan for each resolved year.

We batch local dynamic downloads by point and shared static or annual downloads by tile, or by tile-year pair. Each point worker requests all multiband point-centered exports needed for
that point. Each tile worker requests one shared tile for one layer, or one shared tile for one layer-year pair. This batching strategy makes the download process parallel, resumable, and easy to deduplicate.

\paragraph{Handling large exports.}
Before we request a GeoTIFF from Earth Engine, we estimate its size from the requested bounds, meters-per-pixel scale, number of bands, and float32 storage cost. We compare this estimate to
Earth Engine's practical direct-download limit of about 50 MB. If the estimate is near this limit, we subdivide the request into temporary subtiles, download those subtiles separately, and mosaic them locally into one final raster. We use this
procedure for both point-centered chips and shared study-level tiles. We write masked pixels to disk as a dedicated nodata sentinel value (\(-9999\)) so that missingness survives the
GeoTIFF export unchanged.

\paragraph{Reading windows and aligning layers.}
Raw downloads from different sources do not share a common grid, CRS, or resolution, so we align them only when we build model inputs. For shared tiles, we identify all stored tiles whose
bounding boxes intersect the requested lat-lon window and merge them on demand. 
For raster-based datasets, we reproject all layers for each sample onto a shared grid in a WGS84 azimuthal equidistant (AEQD) projection centered on the step's starting location. 
We warp every other layer onto the
same extent, affine transform, and output array shape.
We use bilinear interpolation for continuous layers and nearest-neighbor interpolation for categorical layers. 

After reprojection, 
we convert the \(-9999\) nodata sentinel back to \texttt{NaN} immediately after reading and construct a boolean validity mask so downstream models can distinguish missing pixels from observed zeros.

Some derived layers depend on spatial gradients and therefore require nontrivial spatial support. We enforce a minimum read window of \(2 \times 2\) pixels for this reason. We derive slope,
aspect, and terrain ruggedness from DEM after loading. We derive orographic uplift from slope, aspect, and ERA5-Land hourly wind. We derive thermal uplift from ERA5-Land hourly surface
variables together with ERA5 hourly atmospheric variables. We derive spectral indices from the Landsat bands after loading rather than treating them as
independent raw downloads.

\paragraph{Model-specific data preparation and storage.}

We store raster data in three layers.

First, we store a raw raster cache for each timescale. This cache contains multiband point-centered chips for local dynamic predictors, reusable shared tiles for static and annual
predictors, one index per shared layer (and one per year for annual layers), and request metadata that records which downloads completed successfully. Because the cache key depends on the
timescale-specific extent and scale, these raw caches are distinct across benchmark horizons even when they reference the same environmental products.

Second, for SSF-style models, we sample each layer at the candidate coordinates and store dense point-feature arrays to reduce computation during training and evaluation. Candidate coordinates are transformed into each raster's CRS for sampling; terrain-derived layers are first computed on a local metric working grid.
For each layer, we store one floating-point array of sampled values and one boolean array of
sampled validity flags. We also store one structured shared array containing candidate coordinates, timestamps, step lengths, turn angles, individual identifiers, burst identifiers, and
point identifiers. We memory-map these arrays during reading so that later stages can load only the needed rows and bands.

Third, for deepSSF, we store fully aligned raster tensors in streaming tar shards to reduce data loading time during training and evaluation, as loading many small GeoTIFF files for every forward pass is otherwise extremely slow. Each sample contains the spatial tensor, scalar time features, start bearing, target grid, and the start-point identifier. 

\section{Metrics}
\label{sec:appendix-metrics}

Here we provide some additional information about the energy score (\cref{sec:metrics}) and its derivation. 
Recall that, for a given trajectory step $(\ell, t)$ 
and forecast distribution $P_{t,\Delta}$~(see \cref{sec:setup}), the energy score evaluated against \(\ell_{t+\Delta}\), the ground truth location at time $t+\Delta$, is:

\[
\operatorname{ES}\!\left(P_{t,\Delta},\,\ell_{t+\Delta}^{\mathrm{obs}}\right)
=
\underbrace{
\mathbb{E}_{X\sim P_{t,\Delta}}
\bigl\|X-\ell_{t_n+\Delta}^{\mathrm{obs}}\bigr\|_2
}_{\text{expected $L_2$ error of forecast}}
-
\frac{1}{2}
\underbrace{
\mathbb{E}_{X,X'\sim P_{t,\Delta}}
\|X-X'\|_2
}_{\text{spread of forecast}}.
\]


To see why ES takes the form that it does, consider the \textit{energy distance}~\cite{rizzo2016energy} between a predicted distribution $P$ and a true distribution $Q$: $D^2(P,Q) = 2 \mathbb{E}\| X - Y \| - \mathbb{E} \| X - X' \| - \mathbb{E} \|Y - Y'\|$, with $X, X' \sim P$ and $Y, Y' \sim Q.$ If we had access to the target distribution $Q$ (\textit{i.e.}, the true conditional distribution of an animal's future location), we could compute and try to minimize the energy distance between $Q$ and our predicted distribution $P$. However, for any given forecast instance we only observe a single realization $y$, which we treat as a draw from $Q$. In expectation, the energy score satisfies $\mathbb{E}_{Y \sim Q} \operatorname{ES}(P,Y) = \tfrac{1}{2} D^2(P,Q) + \tfrac{1}{2} \mathbb{E} \| Y - Y' \|$, \textit{i.e.}, in expectation over $Q$ the energy score is the energy distance between $P$ and $Q$ up to a term that does not depend on $P$. Therefore, minimizing the expected energy score is equivalent to minimizing the energy distance.

\section{Method details}
\label{sec:appendix-methods}

\textbf{Integrated step-selection functions (iSSF).} iSSFs~\cite{avgar2016integrated} and their extensions model the probability of an animal's next step as a function of both movement features (\textit{e.g.} velocity) and environmental features (see \cref{sec:covariates}). In practice this is implemented as a conditional logistic regression. At training time, a set of future candidates $C_{t,\Delta} = \{\ell_{t+\Delta},t+\Delta\}_{i=1}^{n_{cand}} \in \mathcal{X}$ is sampled from parametric distributions fit to the training data. A model then predicts movement probabilities $P_{t,\Delta}$ over the discrete set of candidates $C_{t,\Delta}$.

In our experiments, we benchmark two iSSF baselines: \textbf{(1) iSSF Linear} uses a simple linear layer as the prediction function, matching prior work. Depending on the experiment, inputs may include movement-derived, time-derived, and environmental covariates, as well as linear ``interactions'' between them. 
\textbf{(2) iSSF MLP} uses a multilayer perceptron (MLP) in place of the linear layer used by standard SSFs. We include the same set of covariates, but no interaction terms, instead allowing the MLP to learn any linear or nonlinear interactions between input variables.

\textbf{MoveFormer.} MoveFormer~\cite{cifka2023moveformer} utilizes the same candidate-scoring framework as iSSFs, but incorporates several attention-based~\cite{vaswani2017attention} components. First, it adds additional trajectory history before the most recent step by summarizing past steps using a transformer encoder, with one token per historical step. It then embeds the current set of candidates, performs dot-product attention between the historical embedding and each candidate embedding, and outputs candidate selection probabilities over the same discrete support as iSSFs. MoveFormer also includes token representing the taxonomy of the species of interest, and is the only baseline designed to be trained across studies and across species.

\textbf{deepSSF.} deepSSF~\cite{forrest2026predicting} replaces the candidate scoring of iSSFs with raster inputs processed by (1)~a convolutional habitat-selection head, which captures spatial structure over the range of possible next steps, and (2)~a movement head with convolutional, pooling, and fully connected layers that predicts the parameters of a step-length and turn-angle movement kernel. The resulting habitat-selection and movement surfaces are then combined to produce a predicted next-step probability surface.


\subsection{Implementation}
\label{sec:appendix-impl}

We implement iSSFs in PyTorch. For each valid step start, we construct one discrete choice set containing
\(n_{\text{cand}}\) sampled candidates plus the observed endpoint, so each training example has \(P=n_{\text{cand}}+1\) alternatives. We fit the candidate proposal distributions on the
training split only: step lengths are sampled from a Gamma distribution fit to valid training-step lengths (with location fixed at zero), and turn angles are sampled from a von Mises
distribution fit to valid training turn angles, matching prior work~\cite{avgar2016integrated}. If either parametric fit fails, we fall back to empirical resampling from the corresponding training distribution. We seed candidate
generation deterministically from a global random seed and the step identifier, which makes the sampled choice sets exactly reproducible. We append the observed endpoint as the final alternative in each choice set.

For each alternative, we compute three movement features---step length, log step length, and \(\cos(\text{turn angle})\)---and we sample all environmental covariates at the candidate
endpoint. We standardize environmental covariates with training-split means and standard deviations, and we standardize the three movement features using the corresponding training-split
moments. When time covariates are enabled, we encode time of year and time of day as four cyclic features \((\sin,\cos)\); when individual identity is enabled, we include an indexed
individual-specific effect. 

For \textbf{iSSF Linear}, we use a linear utility function with three groups of terms: (i) one coefficient for each environmental covariate at the candidate endpoint, (ii) one coefficient
for each movement feature (step length, log step length, and \(\cos(\text{turn angle})\)), and (iii) one coefficient for each pairwise product between a movement feature and an
environmental covariate. When time covariates are present, we also include time main effects
together with time-by-environment and time-by-movement interactions. The resulting choice probability is
\[
\Pr(y_i=p \mid C_i)=\frac{\exp(\eta_{ip})}{\sum_{j=1}^{P}\exp(\eta_{ij})}.
\]
We train this model with cross-entropy loss over the \(P\) alternatives in each choice set, taking the observed endpoint as the target class. This is algebraically the same likelihood as
conditional logistic regression. 
Standard R implementations typically fit the same model by reshaping the data into one row per alternative, defining one stratum per step, and optimizing the equivalent conditional logit /
Cox partial likelihood with \texttt{clogit}. Our implementation differs in data layout and optimizer. We use tensorized batches of whole choice sets and optimize the same conditional
likelihood with Adam, which lets us share one training and evaluation pipeline across the PyTorch baselines and scale to repeated large-benchmark runs.

For \textbf{iSSF MLP}, we keep the same candidate-generation procedure, feature construction, and conditional choice loss, but we replace the linear score \(\eta_{ip}\) with an MLP applied
candidate-wise to the concatenated feature vector. In this variant, we do not manually specify interaction terms. Instead, the network receives the base movement, environmental, temporal,
and optional individual-level inputs and learns linear and nonlinear interactions directly. We train both iSSF variants with the same PyTorch training scaffold, using minibatches of choice sets, Adam optimization, checkpointing, and early stopping on held-out negative log-likelihood.

\section{Hyperparameters}
\label{appendix:hparams}

\textbf{Covariates.} 
For all experiments we include some experiment-specific set of the environmental covariates in \cref{tab:movebench_covariates_summary}, as well as movement and time-derived covariates for all methods, which have been shown to improve performance in prior work~\cite{avgar2016integrated,forrest2025predicting,forrest2026predicting}. 
First, for two locations $\ell$ and $\ell'$, we define a \emph{step} beginning at $\ell$ and ending at $\ell'$ with displacement vector $\delta \ell'= \ell'-\ell$. Then, the corresponding \emph{step length} is $d' = \|\delta \ell'\|_2$. When \(d'>0\), we define the corresponding \emph{bearing} as the angle
$\theta' = \operatorname{atan2}\!\bigl((\delta \ell')_2,(\delta \ell')_1\bigr) \in (-\pi,\pi]$.
Finally the \emph{turn angle} is the signed angular increment between successive bearings,
$\alpha' = \operatorname{wrap}\!\bigl(\theta'-\theta\bigr) \in (-\pi,\pi]$.

We then include as covariates for each observed and candidate step: (1)~step length, (2)~log step length, (3)~cosine of turn angle, (4-5)~sine and cosine of normalized day-of-year, and (6-7) sine and cosine of normalized time-of-day~\cite{avgar2016integrated,forrest2025predicting}.  For iSSF Linear, we include linear interaction terms between all movement- and time-derived variables and all environmental variables, following prior work. That is, for each movement- or time-derived variable and each environmental variable, we construct an additional feature equal to their product. This results in $(7+1) \times 160 = 1280$ input features for iSSF Linear and $7 + 160 = 167$ input features for the other methods.
We z-score normalize environmental and movement covariates using training set statistics.

\textbf{Candidate selection.} To be consistent with prior work, we fit a gamma distribution for step-lengths and a Von Mises distribution to turn-angles for all experiments. We use $n_{cand}= 100$ candidates for all steps for all experiments.

\textbf{Time horizons.} We evaluate forecasting methods at a fixed set of forecast horizons $\Delta \in \{\text{1 hour}, \text{12 hours}, \text{1 day}, \text{7 days}\}$. For each horizon $\Delta$, we evaluate \cref{eq:es} at every test-set observation $(\ell_n,t_n)$ for which a valid target location is available within a tolerance of time $t_n+\Delta$. We call this set of observations $\tau_\Delta \subseteq \tau$. We then average these scores over forecast origins for each individual to obtain a per-individual per-timescale mean score $\operatorname{ES}_\Delta = \frac{1}{|\tau_\Delta|}\ \sum_{i=1}^{|\tau_\Delta|} \operatorname{ES} (P_{t_i,\Delta}, \ell_{t_i+\Delta})$. For our experiments, we set this tolerance to $\pm \Delta/3$. 

There are a small number of studies that have no valid test set observations for a given timescale due to either no matching end steps within this tolerance or a native sample rate larger than the specified timescale (\textit{e.g.} studies that were originally sampled with fix rates longer than one hour). During evaluation, we simply skip these study-timescale pairs. Any relevant aggregations (\textit{e.g.} \cref{fig:results}) do not consider skipped study-timescale pairs in their mean and variance calculations.

\textbf{Evaluation extent.} To standardize evaluation, we generate predictions for each method over the entire $101 \times101$ pixel raster of every test point at every time horizon, and compute energy scores on this $101 \times 101$ discrete domain. For methods that are trained with candidate points sampled from parametric distributions (\textit{i.e.} iSSF and MoveFormer), we first compute the renormalized probability assigned to each raster cell under the fitted step-length and turn-angle distributions and then combine this with the model output for each cell to obtain the final predictive surface. This corrects for the mismatch between training on non-uniformly sampled candidate locations and evaluating on a fixed uniform grid.

\textbf{Hyperparameters.} We first perform a small hyperparameter sweep of all methods on \movebench{}-Mini for fair comparison. 
For iSSF linear, we swept over two batch sizes and three learning rates; for iSSF MLP, we swept over batch size, learning rate, number of hidden layers, and hidden layer size; for deepSSF, we swept over learning rate, output channels, and hidden dimension; for MoveFormer, we swept over learning rate, number of transformer layers, number of transformer heads, and context length. We used the best setting from each method for all of our main results. 
We train the iSSF methods on CPU and deepSSF and MoveFormer on a GPU cluster.
We use the temporal hold-out sets as validation during training. We evaluate methods during training roughly every $10,000$ training samples and perform early stopping if validation loss has not decreased in $10$ evaluation runs. We then evaluate the best checkpoint according to validation loss on both test sets.

\clearpage

\section{Additional experiment details}
\label{sec:appendix-results}





\subsection{All experimental results}

\begingroup
\footnotesize
\setlength{\tabcolsep}{1.5pt}
\setlength{\LTleft}{0pt}
\setlength{\LTright}{0pt}
\renewcommand{\arraystretch}{1.05}
\newcolumntype{L}[1]{>{\RaggedRight\arraybackslash}p{#1}}
\newcolumntype{C}[1]{>{\Centering\arraybackslash}p{#1}}
\begin{longtable}{@{}L{0.16\textwidth}L{0.10\textwidth}*{8}{C{0.078\textwidth}}@{}}
\caption{Raw energy scores for each study and method. Lower is better. Dashes indicate unavailable results. MoveFormer denotes the shared model.}\label{tab:final_two_row_energy_scores}\\
\toprule
\multirow{2}{*}{Study} & \multirow{2}{*}{Method} & \multicolumn{4}{c}{ES, Temporal split $\downarrow$} & \multicolumn{4}{c}{ES, Individual split $\downarrow$} \\
\cmidrule(lr){3-6}\cmidrule(lr){7-10}
& & 1h & 12h & 1d & 7d & 1h & 12h & 1d & 7d \\
\midrule
\endfirsthead
\caption[]{Energy score results (continued).}\\
\toprule
\multirow{2}{*}{Study} & \multirow{2}{*}{Method} & \multicolumn{4}{c}{ES, Temporal split $\downarrow$} & \multicolumn{4}{c}{ES, Individual split $\downarrow$} \\
\cmidrule(lr){3-6}\cmidrule(lr){7-10}
& & 1h & 12h & 1d & 7d & 1h & 12h & 1d & 7d \\
\midrule
\endhead
\midrule
\multicolumn{10}{r}{\emph{Continued on next page}}\\
\endfoot
\bottomrule
\endlastfoot
\multirow{6}{=}{\citet{signer2019AnimalMovementTools}} & don't move & -- & 686 & 878 & 1482 & -- & -- & -- & -- \\
 & crw & -- & 538 & 682 & 1100 & -- & -- & -- & -- \\
 & issf linear & -- & 537 & 668 & 1071 & -- & -- & -- & -- \\
 & issf mlp & -- & 522 & 662 & 1056 & -- & -- & -- & -- \\
 & deepssf & -- & 529 & 664 & 1153 & -- & -- & -- & -- \\
 & moveformer & -- & 496 & 627 & 1124 & -- & -- & -- & -- \\
\midrule
\multirow{6}{=}{\citet{accio2022TimingIsCritical}} & don't move & 14709 & 102696 & 185534 & 1150541 & 15027 & 114012 & 188756 & 1098839 \\
 & crw & 11345 & 79685 & 130827 & 776299 & 11635 & 88671 & 134543 & 735285 \\
\end{longtable}
\endgroup

\subsection{Per-study MoveFormer}

\begin{figure}[h!]
    \centering
    \includegraphics[width=\linewidth]{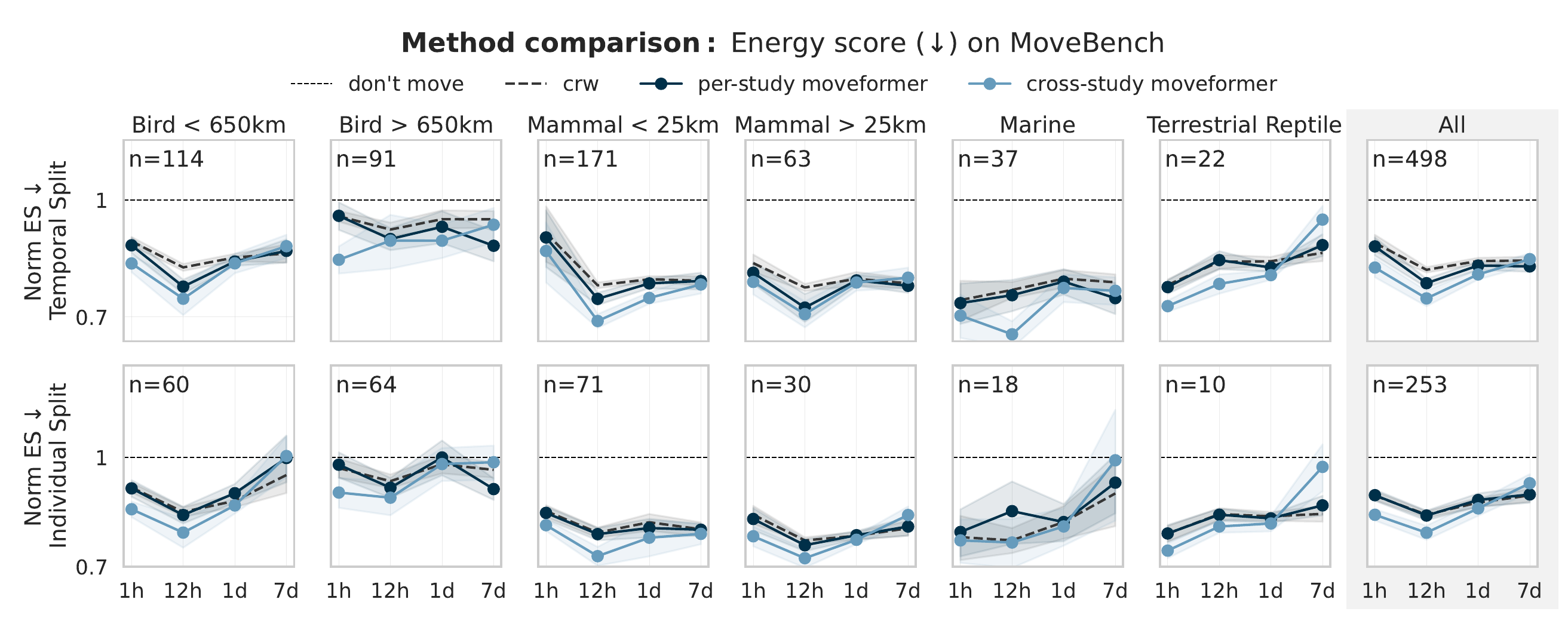}
    \caption{Training one MoveFormer model per study per time horizon vs. training one MoveFormer model across all studies for each time horizon. The latter is the published default that is used in our main experiments.}
    \label{fig:moveformer}
\end{figure}

MoveFormer is the only method we benchmarked that is designed to train across studies. For comparison with the other methods, which train one model per-study, we also experimented with per-study MoveFormer training. On average, cross-study training was beneficial. See \cref{fig:moveformer}.

\subsection{Compute and storage requirements}
\label{sec:appendix-compute}

We trained 3,360 study-specific models across 147 studies, each evaluating one method at one timescale, and four shared MoveFormer training runs that were trained across all studies at each timescale as in the original implementation.
We ran  experiments on a heterogeneous SLURM cluster. 
All jobs were allocated 16 CPU cores and 64 GB RAM, however this is more than required for most experiments. Baseline and iSSF variants ran CPU-only, whereas deepssf and moveformer used a single GPU on either A100, V100, L40S, H100, H200, RTX3090/3080, A6000, or RTX6000 Ada nodes. Unless noted otherwise, the numbers below are median wall-clock times over completed tasks. Note that jobs were assigned to GPU nodes randomly, and run time is affected by the size of the study, so the times below may not be indicative of run time on a particular GPU generally.


\begin{itemize}
      \item The ``don't move'' baseline took 1.1~min, and the correlated random walk baseline took 1.9~min.
      \item Linear iSSF took 6.9~min, and MLP iSSF took 11.2~min.
      \item deepSSF took 11.0~min overall. Medians by GPU were 8.4~min on L40S, 9.7~min on H200, 10.7~min on H100, 12.1~min on A100, 13.1~min on A6000,
      and 17.3~min on V100.
      \item Training the four shared MoveFormer models took 7.3~h (1~h timescale), 13.4~h (12~h), 5.1~h (1~d), and 6.0~h (7~d), each on one A100, for a
      total of 31.7 GPU-hours. Subsequent per-study scoring took 4.6~min overall: 4.7~min on A100 and 4.4~min on L40S.
      \item The 90th-percentile runtimes were 28.5~min for linear iSSF, 36.0~min for MLP iSSF, 107.1~min for deepSSF, and 55.7~min for MoveFormer
      scoring.
  \end{itemize}

The full dataset, including all covariates for all timescales for all studies, totals 7 terabytes using 32-bit floating point representations. We foresee that most users of the benchmark will not require all of this data. For instance, they main train models with a subset of covariates and/or for a subset of studies. Our Python package enables download of arbitrary subsets of covariates, studies, and timescales directly from a hosted archive. For covariates or timescales not included in our experiments, the package also enables downloading arbitrary layers, timescales, and studies (including custom studies that a user can import themselves), via Google Earth Engine. For reference, all covariates for all timescales of \movebench-Mini totals approximately 500 gigabytes, and the ``lite'' set of covariates (\cref{sec:covariates}) for all timescales of \movebench-Mini totals approximately 20 gigabytes. Quantized versions of all covariates will also be released for additional space savings.



\end{document}